\documentclass[10pt,twocolumn,letterpaper]{article}

\usepackage{iccv}
\usepackage{times}
\usepackage{epsfig}
\usepackage{graphicx}
\usepackage{amsmath}
\usepackage{amssymb}
\usepackage{caption}
\usepackage{graphicx}
\usepackage{amsmath}
\usepackage{amssymb}
\usepackage{booktabs}
\usepackage[utf8]{inputenc} 
\usepackage[T1]{fontenc}    
\usepackage{url}            
\usepackage{booktabs}       
\usepackage{amsfonts}       
\usepackage{nicefrac}       
\usepackage{microtype}      
\usepackage{xcolor}         
\usepackage[linesnumbered,ruled,vlined]{algorithm2e}
\usepackage{array}
\usepackage{amssymb}
\usepackage{latexsym}
\usepackage{amsmath,amssymb}
\usepackage{graphicx}
\usepackage{multirow}
\usepackage{adjustbox}
\usepackage{makecell}
\usepackage[nottoc]{tocbibind}\usepackage{url}
\usepackage{wrapfig}

\SetCommentSty{mycommfont}
\newlength{\commentWidth}
\usepackage{amsmath,amsfonts,bm}

\def\eqref#1{equation~\ref{#1}}

\def\1{\bm{1}}

\DeclareMathAlphabet{\mathsfit}{\encodingdefault}{\sfdefault}{m}{sl}
\SetMathAlphabet{\mathsfit}{bold}{\encodingdefault}{\sfdefault}{bx}{n}

\newcommand{\cb}{{\boldsymbol c}}
\newcommand{\db}{{\boldsymbol d}}

\newcommand{\qb}{{\boldsymbol q}}

\newcommand{\xb}{{\boldsymbol x}}

\newcommand{\zb}{{\boldsymbol z}}

\newcommand{\beq}{\begin{equation}}
\newcommand{\eeq}{\end{equation}}
\newcommand{\beqa}{\begin{eqnarray}}
\newcommand{\eeqa}{\end{eqnarray}}

\newcommand{\epsilonb}{\boldsymbol{\epsilon}}

\usepackage[breaklinks=true,bookmarks=false]{hyperref}
\hypersetup{
    colorlinks=true,
    linkcolor=red,
    filecolor=magenta,      
    urlcolor=magenta,
}

\usepackage[capitalize]{cleveref}
\iccvfinalcopy 

\def\iccvPaperID{XXXX} 

\begin{document}


\title{PODIA-3D: Domain Adaptation of 3D Generative Model Across \\ Large Domain Gap Using Pose-Preserved Text-to-Image Diffusion}

\author{Gwanghyun Kim$^1$ \qquad Ji Ha Jang$^1$ \qquad \stepcounter{footnote} Se Young Chun$^{1,2}$\thanks{} \ \\
$^1$Dept. of Electrical and Computer Engineering, $^2$INMC \&  IPAI \\
Seoul National University, Republic of Korea \\
{\tt\small \{gwang.kim, jeeit17, sychun\}@snu.ac.kr}
}

\ificcvfinal\thispagestyle{empty}\fi

\twocolumn[{%
\renewcommand\twocolumn[1][]{#1}%
\maketitle
\begin{center}
    \centering
    \captionsetup{type=figure}
    \vspace{-3em}
    \includegraphics[width=\textwidth]{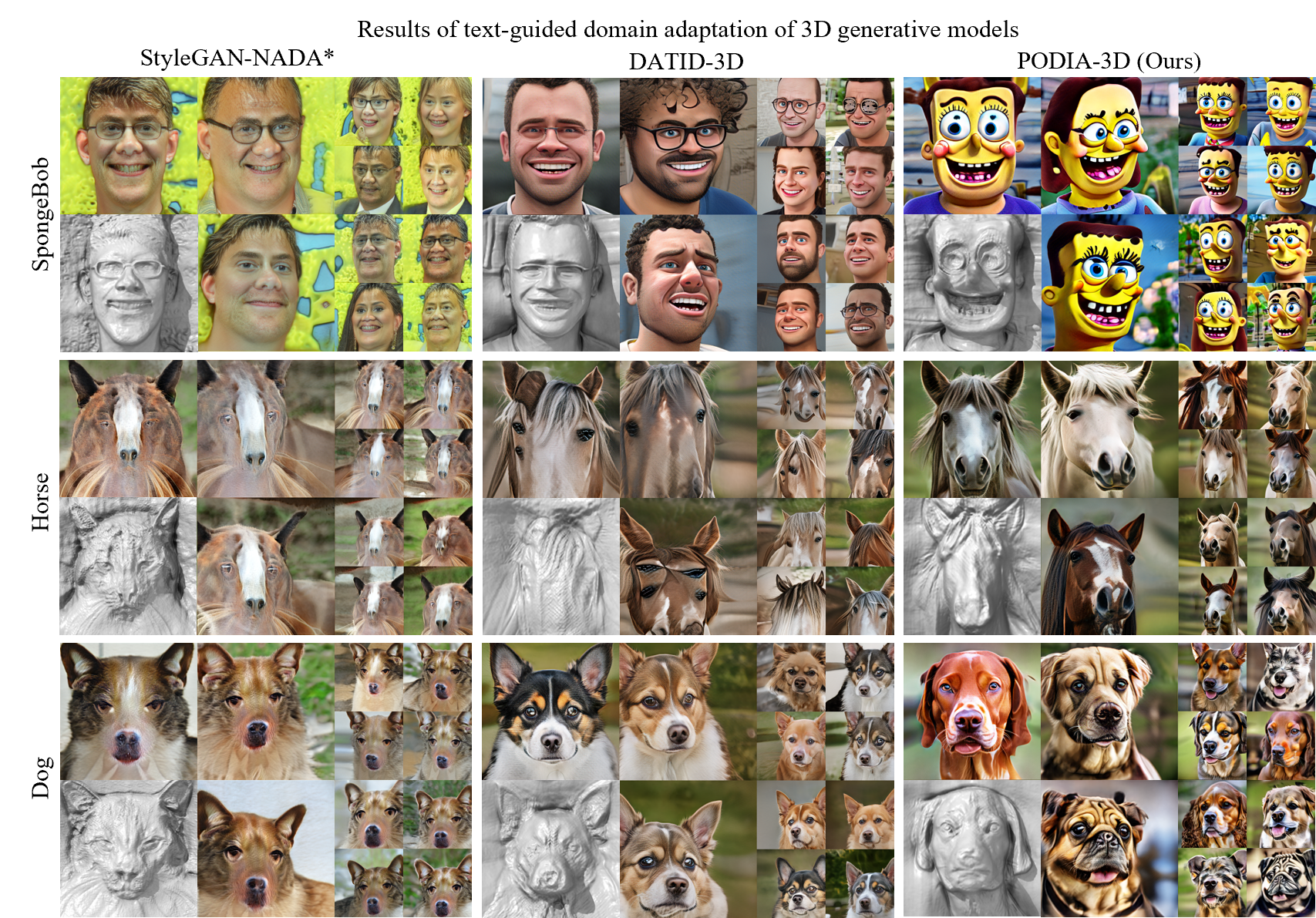}
    \vspace{-1.5em}
    \captionof{figure}{
    Our PODIA-3D successfully adapts 3D generators across significant domain gaps, producing excellent text-image correspondence and 3D shapes, while the baselines fail. See the supplementary videos at \href{https://gwang-kim.github.io/podia_3d}{\small{\texttt{gwang-kim.github.io/podia\_3d}}}.}
    \label{fig1}
\end{center}%
}]
{
  \renewcommand{\thefootnote}%
    {\fnsymbol{footnote}}
  \footnotetext[2]{Corresponding author.}
}

\begin{abstract}
Recently, significant advancements have been made in 3D generative models, however training these models across diverse domains is challenging and requires an huge amount of training data and knowledge of pose distribution. Text-guided domain adaptation methods have allowed the generator to be adapted to the target domains using text prompts, thereby obviating the need for assembling numerous data. Recently, DATID-3D presents impressive quality of samples in text-guided domain, preserving diversity in text by leveraging text-to-image diffusion. However, adapting 3D generators to domains with significant domain gaps from the source domain still remains challenging due to issues in current text-to-image diffusion models as following: 1) shape-pose trade-off in diffusion-based translation, 2) pose bias, and 3) instance bias in the target domain, resulting in inferior 3D shapes, low text-image correspondence, and low intra-domain diversity in the generated samples. To address these issues, we propose a novel pipeline called PODIA-3D, which uses pose-preserved text-to-image diffusion-based domain adaptation for 3D generative models. We construct a pose-preserved text-to-image diffusion model that allows the use of extremely high-level noise for significant domain changes. We also propose specialized-to-general sampling strategies to improve the details of the generated samples. Moreover, to overcome the instance bias, we introduce a text-guided debiasing method that improves intra-domain diversity. Consequently, our method successfully adapts 3D generators across significant domain gaps. Our qualitative results and user study demonstrates that our approach outperforms existing 3D text-guided domain adaptation methods in terms of text-image correspondence, realism, diversity of rendered images, and sense of depth of 3D shapes in the generated samples.

    \end{abstract}

\vspace{-1em}

\section{Introduction}
\label{sec1_introduction}

\begin{figure*}[!t]
    \centering
    \includegraphics[width=\linewidth]{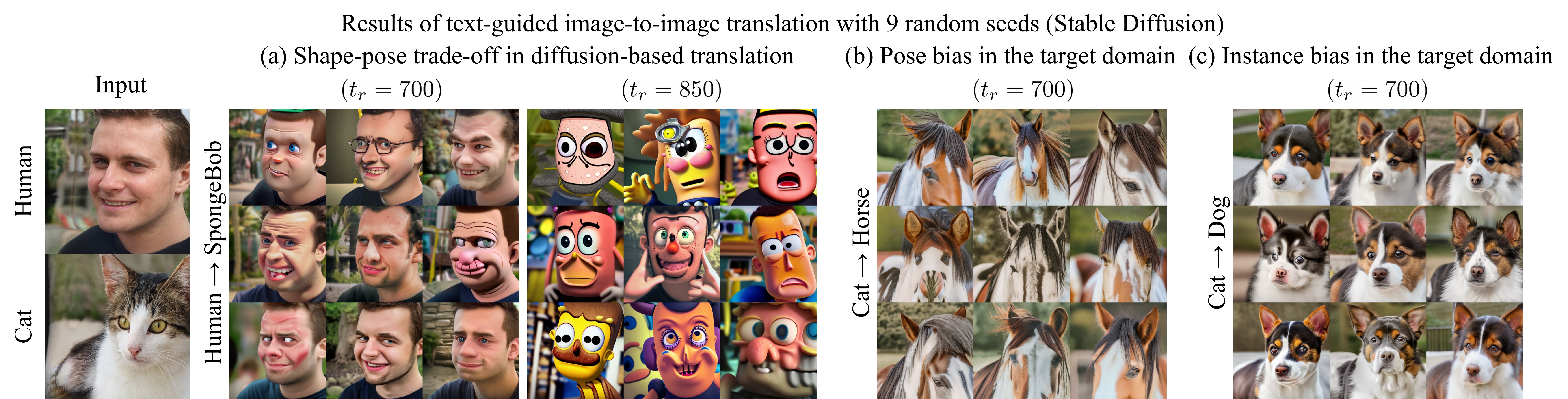}
    \vspace{-2em}
    \caption{Issues in pose-aware target generation for domain adaption of 3D generative models using current text-to-image diffusion models: (a) a shape-pose trade-off in diffusion-based translation, (b) pose bias, and (c) instance bias in the target domain.}
    \vspace{-1.em}
    \label{fig2}
\end{figure*}

Recently, 3D generative models ~\cite{liao2020towards, szabo2019unsupervised, gadelha20173d, henzler2019escaping, nguyen2019hologan, nguyen2020blockgan, zhu2018visual, wu2016learning, chan2021pi, nguyen2019hologan,niemeyer2021giraffe, schwarz2020graf, gu2022stylenerf, zhou2021cips, chan2022efficient} have been advanced to enable multi-view consistent and explicitly pose-controlled image synthesis. 
However, training state-of-the-art 3D generative models is challenging due to the requirement of a large number of images and knowledge about their camera pose distribution. This prerequisite has resulted in limited applications of these models to only a few domains.

Text-guided domain adaptation methods such as StyleGAN-NADA~\cite{gal2021stylegan}, HyperDomainNet~\cite{alanov2022hyperdomainnet}, DATID-3D~\cite{kim2022datid}, and StyleGANFusion~\cite{song2022diffusion} have emerged as a promising solution to overcome the challenge of need for additional data of the target domain. These methods leverage CLIP~\cite{radford2021learning} or text-to-image diffusion models~\cite{rombach2022high, ramesh2022hierarchical, saharia2022photorealistic} that are pretrained on a large number of image-text pairs.

Although non-adversarial fine-tuning methods like StyleGAN-NADA~\cite{gal2021stylegan}, HyperDomainNet~\cite{alanov2022hyperdomainnet}, and StyleGANFusion~\cite{song2022diffusion} have demonstrated impressive results, they suffer from the inherent loss of diversity in a text prompt and suboptimal text-image correspondence, as illustrated in Fig.~\ref{fig1} (See results of StyleGAN-NADA*).
Recently, a diversity-preserved domain adaptation method called DATID-3D~\cite{kim2022datid} has been developed for 3D generators, which achieves compelling quality of multi-view consistent image synthesis in text-guided domains. This method generates pose-aware target dataset using text-to-image diffusion models and fine-tunes the 3D generator on the target dataset.

Despite the use of this method, the adaptation of 3D generators to domains that have significant domain gaps from the source domain remains challenging due to the problems encountered in the current text-to-image diffusion models. 
1) shape-pose trade-off in diffusion-based translation: For text-guided pose-aware target generation, we first perturb the source image or latent $\xb^{\text{src}}_0$ until $t_r \in [1,T]$ such that $\xb^{\text{trg}}_0$ generated from $\xb^{\text{src}}_{t_r}$ should represent the features corresponded the target domains without altering pose of  $\xb^{\text{src}}_0$.
However, our investigations show that when the target domain requires selecting a high $t_r$ to achieve a significant structural change, preserving the pose is not guaranteed, as depicted in Fig.~\ref{fig2}(a).
Consequently, shifting the generator to a target domain that requires significant shape changes can lead to poor 3D shapes or low text correspondence, as illustrated in Fig.~\ref{fig1} (See SpongeBob by DATID-3D~\cite{kim2022datid}).  
2) We found that a publicly available text-to-image diffusion model, has pose bias issues for certain text prompts representing the target domains, as illustrated in Fig.~\ref{fig2}(b).
Accordingly, the shifted generators guided by these text prompts result in either poor 3D structure as represented in Fig.~\ref{fig1} (See Horse by DATID-3D~\cite{kim2022datid}).  
3) We also found that the text-to-image diffusion models often generate images with one or a few instances among many instances representing the text prompts as represented in Fig.~\ref{fig2}.
In consequence, the shifted generators guided by these text prompts result in low intra-domain diversity as represented in Fig.~\ref{fig1}.  (See Dog by DATID-3D~\cite{kim2022datid}).  

To address these issues, we propose a novel pipeline called PODIA-3D, a method of \textbf{PO}se-preserved text-to-image \textbf{DI}ffusion-based doamin \textbf{A}daptation for \textbf{3D} generative models.
We construct pose-preserved text-to-image diffusion models. 
We first collect target images that have the same pose but different shapes with source images through 3 strategies: 1) identity mixing, 2) text-guided image translation with pose-guaranteed prompts, and 3) utilizing the different domain generator.
Then, we fine-tune the depth-guided diffusion model to make it ignore the shape information from the depth map and focus only on pose information. 
Furthermore, we propose a specialized-to-general sampling strategy to improve details of generated images and resolve the detail bias issue.
Using pose-preserved diffusion models and specialized-to-general sampling, we are able to synthesize pose-consistent target images with excellent text-image correspondence by using extremely high-level noise for large shape change.
 We then fine-tune the state-of-the-art 3D generator adversarially on the generated target images.
Moreover, to improve intra-domain diversity, we propose a text-guided debiasing method, which enables the fine-tuned generator to reach the diverse modes.
As a result, our method effectively adapts 3D generators across significant domain gaps, generating excellent text-image correspondence and 3D shapes, as shown in Fig.~\ref{fig1}. Our approach has been demonstrated to outperform existing 3D text-guided domain adaptation methods in terms of text-image correspondence, realism, diversity of rendered images, and sense of depth of 3D shapes in the generated samples via the qualitative results and user study.

\section{Related Works}

\subsection{3D generative models}
Recent advancements in 3D generative models~\cite{liao2020towards, szabo2019unsupervised, gadelha20173d, henzler2019escaping, nguyen2019hologan, nguyen2020blockgan, zhu2018visual, wu2016learning, chan2021pi, nguyen2019hologan, niemeyer2021giraffe, schwarz2020graf, gu2022stylenerf, zhou2021cips, chan2022efficient} have enabled multi-view consistent and explicitly pose-controlled image synthesis.
Notably, EG3D~\cite{chan2022efficient}, which use StyleGAN2~\cite{karras2020analyzing} generator, in conjunction with neural rendering~\cite{mildenhall2020nerf}, succeed in producing high resolution multi-view consistent images in real-time as well as highly detailed 3D shapes.
However, training modern 3D generative models is more challenging than training 2D generative models, as it requires a significant number of images and detailed information on the camera parameter distribution for those images.

To expand the usability of state-of-the-art 3D generative models to a wider range of domains, including those with significant domain gaps, we introduce PODIA-3D, an approach that employs text-guided adaptation methods for 3D generators using pose-preserved diffusion models to enhance image-text correspondence, 3D shapes, and intra-domain diversity.

\subsection{Text-to-image diffusion models}
Diffusion models have demonstrated great success in the fields of image generation~\cite{ho2020denoising,song2020denoising,song2020score,jolicoeur2020adversarial,dhariwal2021diffusion} and image-text multimodal applications~\cite{rombach2022high, ramesh2022hierarchical, saharia2022photorealistic, kim2022diffusionclip, avrahami2022blended}. In recent years, text-to-image diffusion models trained on large-scale image-text datasets~\cite{rombach2022high, ramesh2022hierarchical, saharia2022photorealistic} have exhibited remarkable performance in generating diverse 2D images from a single text prompt. One variant of text-to-image diffusion models referred to as depth-guided diffusion models~\cite{rombach2022high}, employs depth maps as a conditioning input throughout the generative process to synthesize images that correspond to the provided depth map.

In this work, we propose the pose-preserved text-to-image diffusion model to generate faithful pose-consistent target images to adapt the 3D generator to text-guided domains with large domain gaps.

\subsection{Text-guided domain adaptation}
Domain adaptation methods guided by textual prompts have been developed for 2D generative models, providing a promising solution to the challenge of acquiring additional data for the target domain. These methods utilize CLIP~\cite{radford2021learning} or text-to-image diffusion models~\cite{rombach2022high, ramesh2022hierarchical, saharia2022photorealistic} pretrained on a large number of image-text pairs, allowing for text-driven domain adaptation. StyleGAN-NADA~\cite{gal2021stylegan} and HyperDomainNet~\cite{alanov2022hyperdomainnet} fine-tune pretrained StyleGAN2~\cite{karras2020analyzing} models to shift the domain towards a target domain, utilizing a simple textual prompt guided by CLIP~\cite{radford2021learning} loss.
StyleGANFusion~\cite{song2022diffusion} adopts SDS loss~\cite{poole2022dreamfusion} as guidance of text-guided adaptation of 2D and 3D generators using text-to-image diffusion models.
Although these non-adversarial fine-tuning methods have demonstrated impressive results, they suffer from the inherent loss of diversity in a text prompt and suboptimal text-image correspondence.
DATID-3D~\cite{kim2022datid} achieves impressive quality in multi-view consistent image synthesis for text-guided domains by generating diverse pose-aware target dataset using text-to-image diffusion models and fine-tuning the 3D generator on the target dataset while preserving diversity in the text.
However, adapting 3D generators to domains with significant domain gaps from the source domain using existing methods remains challenging. 
This is because these models suffer from several issues, such as a shape-pose trade-off in diffusion-based translation, pose bias, and instance bias in the target domain. 
As a result, the generated samples often exhibit inferior 3D shapes, low text-image correspondence, and low intra-domain diversity.

To mitigate these issues, we propose a novel pipeline called PODIA-3D, a method of pose-preserved text-to-image diffusion-based domain adaptation for the 3D generative models. 

\begin{figure*}[!t]
    \centering
    \includegraphics[width=\linewidth]{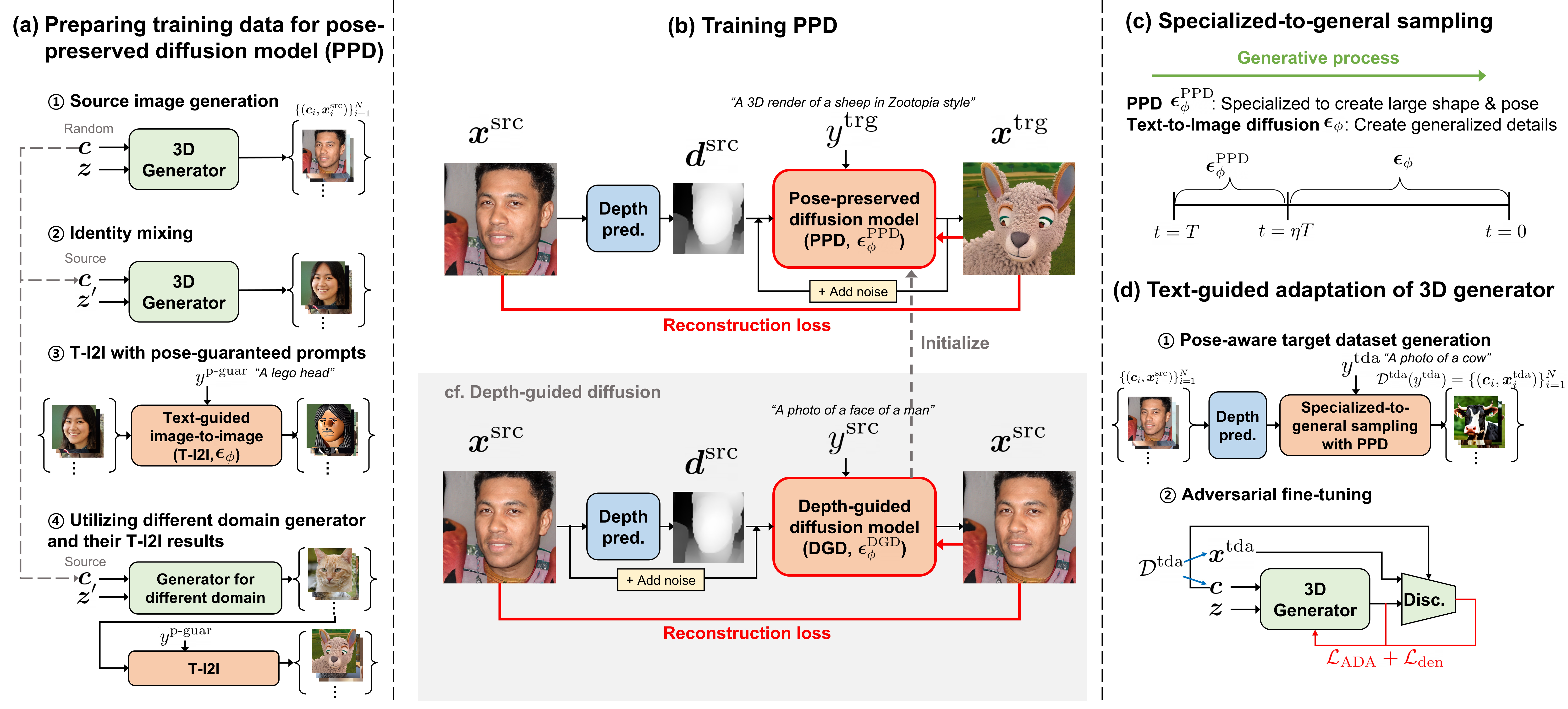}
      \vspace{-1.5em}
    \caption{Overview of PODIA-3D. (a) We prepare data for training pose-preserved diffusion models (PPD) and (b) fine-tune the depth-guided diffusion models on the collected data. (c) We use a specialized-to-general sampling strategies to generate high quality pose-aware target images. (d) Finally, we fine-tune the state-of-the-art 3D generator on them adversarially. }
    \label{fig3}
    \vspace{-1.em}
\end{figure*}

\begin{figure}[!t]
    \centering
    \includegraphics[width=\linewidth]{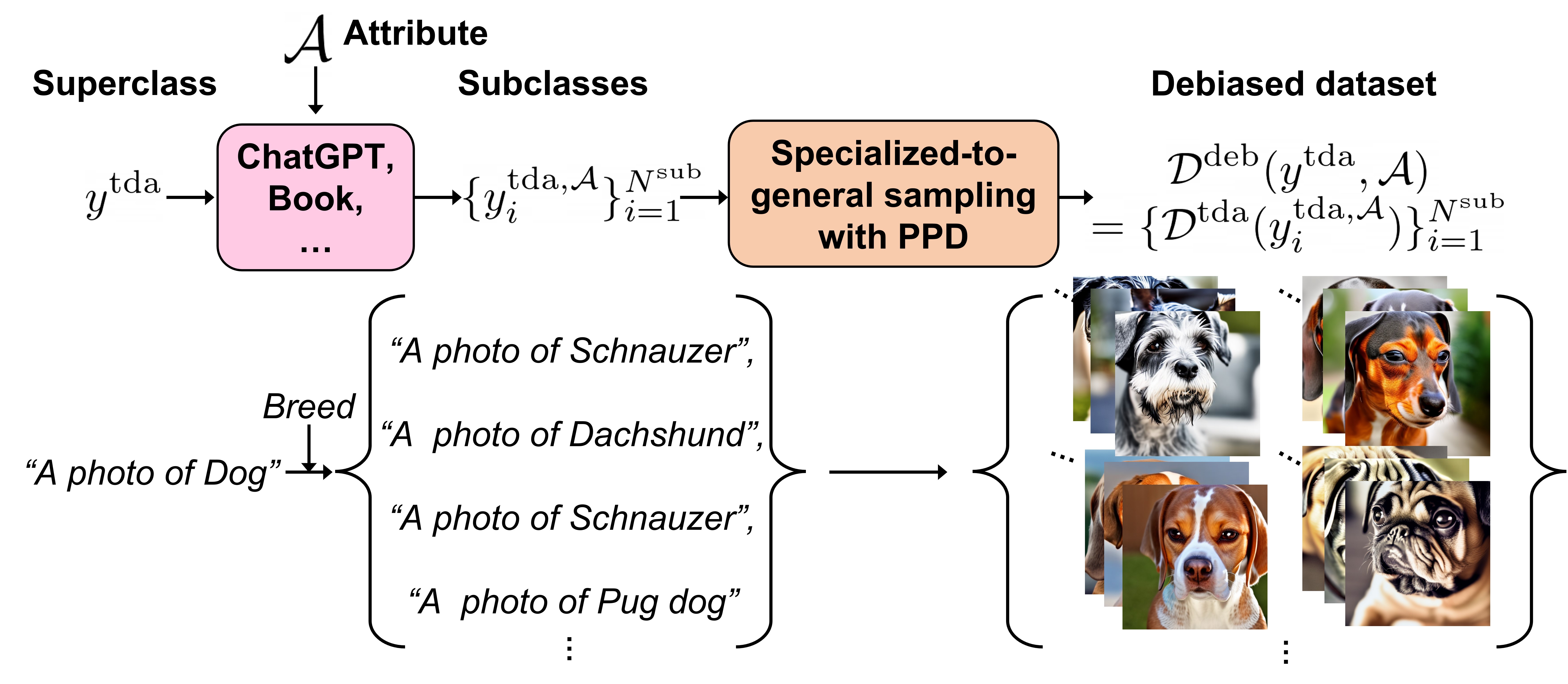}
    \vspace{-2em}
    \caption{Our text-guided debiasing method includes obtaining a set of subclass texts, and then generating a pose-aware target dataset for each subclass text. We combine these datasets to construct a debiased target dataset.
}
    \vspace{-2.em} 
    \label{fig4}
\end{figure}

\begin{figure*}[!t]
    \centering
    \includegraphics[width=\linewidth]{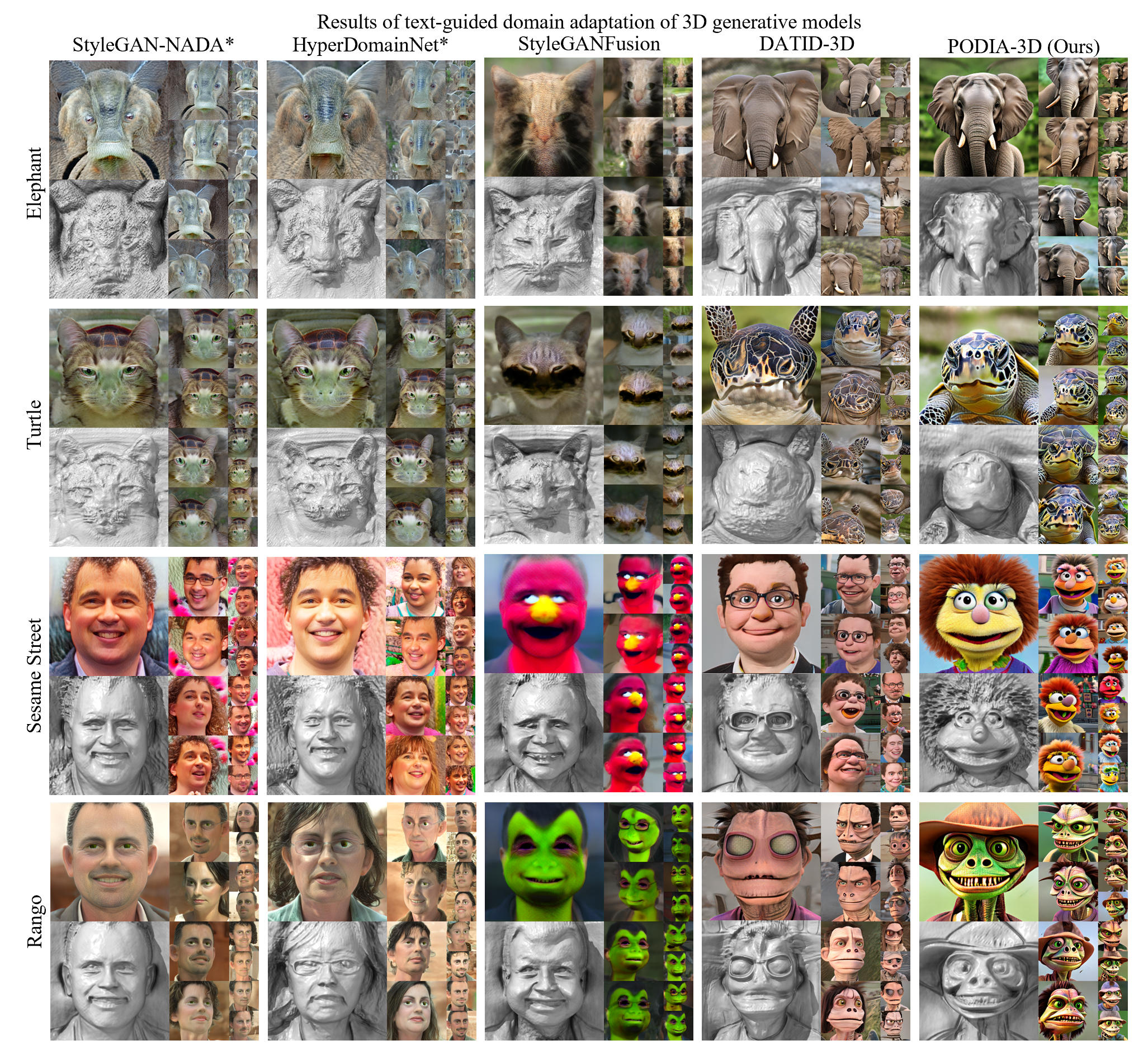}
    \vspace{-2.5em}
    \caption{Qualitative comparison with existing text-guided domain adaptation methods with a star (*) indicating their 3D extensions.  Our method allows to adapt the 3D generative models to domains with huge domain gap, presenting excellent text-image correspondence and 3D shape.}
    \vspace{-1em}
    \label{fig5}
\end{figure*}

 \begin{figure*}[!t]
    \centering
    \includegraphics[width=\linewidth]{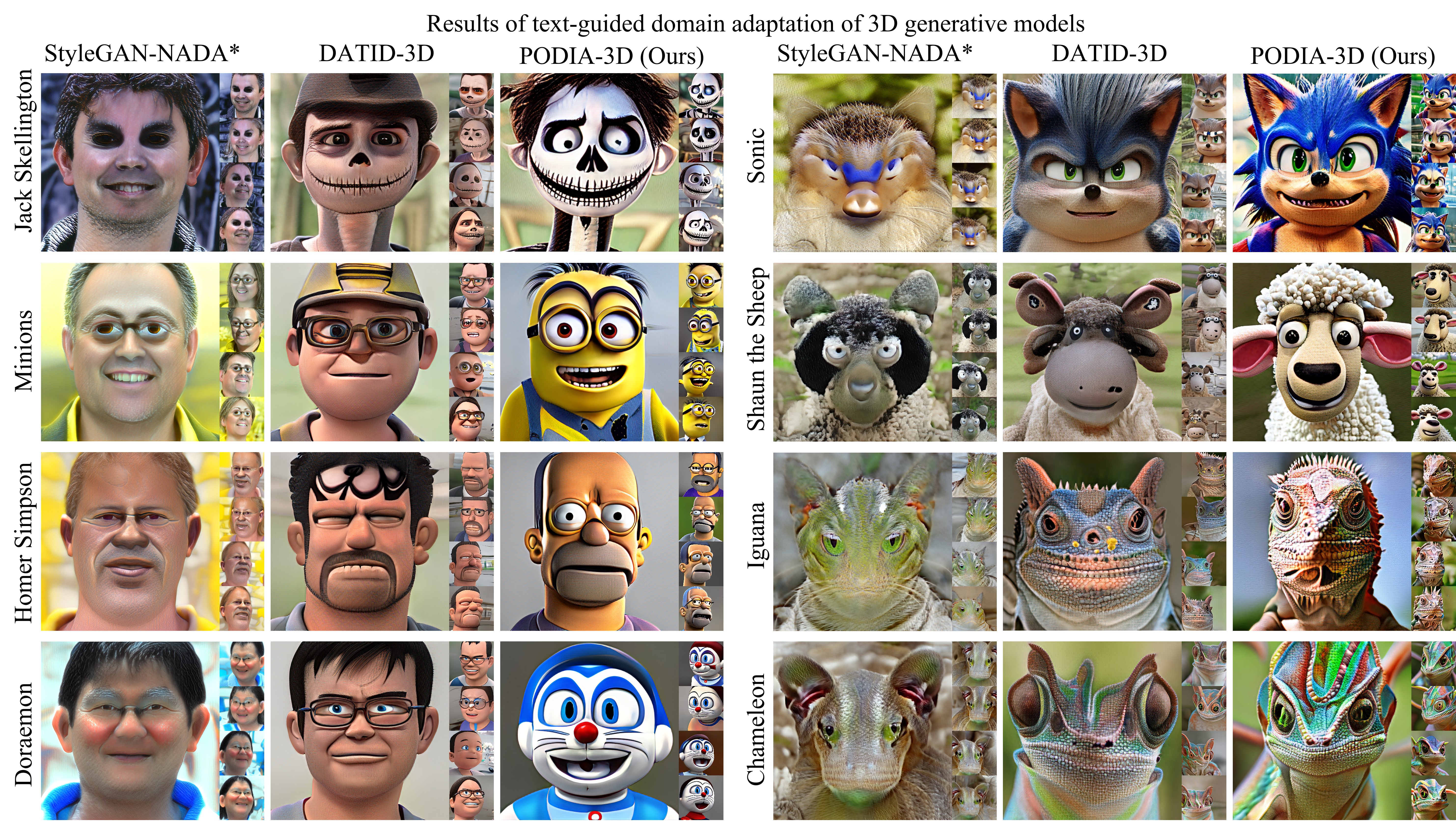}
    \vspace{-2em}
    \caption{Our method succeed in text-guided adaption to the wide range of domains while other baselines show the results with low text-image correspondence. For extended results, see the supplementary Fig.~\ref{fig_supp2} and~\ref{fig_supp3}.}
    \vspace{-0.5em}
    \label{fig6}
\end{figure*}

\section{PODIA-3D}
\label{sec3_method}

We begin by constructing a text-to-image diffusion model that preserves the pose of the source images while generating target images, as shown in Fig.\ref{fig3}(a)(b). We then propose a specialized sampling strategy to enhance the pose-preserved diffusion and improve image details, as demonstrated in Fig.\ref{fig3}(c). Using this approach, we perform pose-preserved diffusion-driven text-guided domain adaptation in two steps: 1) generating a pose-aware target dataset and 2) fine-tuning the 3D generator using adversarial training, as depicted in Fig.\ref{fig3}(d). Furthermore, to address the instance bias observed in certain text prompts, we introduce a text-guided debiasing method, as illustrated in Fig.\ref{fig4}, to improve intra-domain diversity.

\subsection{Pose-preserved text-to-image diffusion models}
We aim to synthesize target images that have faithful pose-consistency, high diversity, and excellent text-image correspondence. To achieve this, we initially consider using the depth-guided diffusion model (DGD)~\cite{rombach2022high}, which generates images conditioned on both depth maps and text prompts, thus producing images consistent with the given depth maps. However, as shown in Fig.~\ref{fig8}, we observed that the strong shape constraints imposed by the depth maps can result in low diversity and poor text-image correspondence, particularly for text prompts that require significant shape changes. To overcome this issue while retaining the benefits of DGD, we develop pose-preserved text-to-image diffusion models (PPD) by fine-tuning DGD to focus only on pose information and ignore shape information during image generation.
\paragraph{Preparation of training data for PPD.}
We prepare training data $\mathcal{D}^{\text{PPD}} = \{(\db_i^{\text{src}}, \qb_i^{\text{trg}},  y^{\text{trg}})\}_{i=1}^{N^{\text{PPD}}}$ for training PPD, which consists of a source depth map $\db_i^{\text{src}}$, the diffusion latent $\qb_i^{\text{trg}}=E^V(\xb_i^{\text{trg}})$ from target image $\xb_i^{\text{trg}}$ encoded by VQGAN encoder $E^V$, and target text prompt $y^{\text{trg}}$ following the process illustrated in Fig.~\ref{fig3}(a).
We start by generating $N^{\text{src}}$ source images $\xb^{\text{src}} = G_{\theta}(\zb, \cb)$ with random latent vectors $\zb$ and camera parameters $\cb$ given the pretrained source 3D generator $G$, which in our case is the EG3D~\cite{chan2022efficient} model trained on $512^2$ FFHQ~\cite{karras2019style} images.
Next, we obtain the source depth maps $\db^{\text{src}}$ using a pretrained depth estimation model.
To collect the set of target images $\xb^{\text{trg}}$ with the same pose as the source images but different shapes, we employ the following three strategies:
1) Identity mixing: We generate images by feeding the same camera parameters $\cb$ with the source images into $G$ but with different latent vectors $\zb$. The prompts for $y^{\text{trg}}$ are chosen to represent the source domain.
2) Text-guided image-to-image translation (T-I2I)~\cite{meng2021sdedit} with pose-guaranteed prompts: We use the pretrained text-to-image model (Stable diffusion~\cite{rombach2022high}) to perform T-I2I on the identity-mixed images for each prompt with guaranteed pose consistency and excellent text-image correspondence, based on our observation. We carefully select the text prompts to avoid overlapping visual features, mitigating bias issues.
3) Using a different domain generator: To achieve further large shape changes, we use the generator trained for a different domain. Specifically, we use the EG3D~\cite{chan2022efficient} model pretrained on AFHQ-cat~\cite{karras2021alias, choi2020stargan} dataset, which is transferred from the FFHQ EG3D model. We generate images to have the same pose as the source images with this model and also use the translated images using T-I2I.

\paragraph{Fine-tuning objective for PPD.}
We fine-tune the copy of pretrained DGD $\epsilon_\phi^{\text{PPD}}$ on $\mathcal{D}^{\text{PPD}}$ using following objective:
\small
\begin{align}
\mathbb{E}_{(\db^{\text{src}}, \qb^{\text{trg}},  y^{\text{trg}}) \in \mathcal{D}^{\text{PPD}}, \epsilonb, t}[\|\epsilonb-\epsilonb^{\text{PPD}}_\phi(\qb^{\text{trg}}_t, y^{\text{trg}}, \db^{\text{src}}, t)\|_2^2], \nonumber
\end{align}
\normalsize
where $\epsilonb \sim \mathcal{N}(0,1),  t \sim \mathcal{U}([1,T])$.

\subsection{Specialized-to-general sampling}
Although PPD trained on augmented data can generate images with corresponding pose and shape to the depth map and text prompt, we discovered the presence of style and detail biases that are inherent in the training data. To address this issue and enhance details, we propose specialized-to-general sampling strategies that leverage the pose-consistent generation capability of the PPD model and the generalization capability of text-to-image diffusion models as presented in Fig.~\ref{fig3}(c). During the first $\eta T$ period, where $\eta \in [0,1]$ is a PPD ratio and $T$ is the number of total diffusion steps, we use the PPD model to generate large structural components and pose information. For the remaining $(1-\eta)T$ period, we utilize Stable diffusion~\cite{rombach2022high}, the general text-to-image diffusion model, to generate small structures or details in the images.

\subsection{Adapting 3D generator to broader domains}
 As illustrated in Fig.~\ref{fig3}(d), we translate the source image $\xb^{\text{src}}$ to yield the target image $\xb^{\text{tda}}$ guided by a text prompt $y^{\text{tda}}$ using PPD and specialized-to-general sampling, constructing the pose-aware target dataset $\mathcal{D}^{\text{tda}}(y^{\text{tda}})=\{(\cb_i, \xb^{\text{tda}}_i)\}^{N}_{i=1}$ . 
Then, we fine-tune 3D generator adversarially using the loss composed of ADA loss $\mathcal{L}_{\text{ADA}}$~\cite{karras2020training} and density regularization loss $\mathcal{L}_{\text{den}}$, following EG3D~\cite{chan2022efficient} and DATID-3D~\cite{kim2022datid}.

PPD and specialized-to-general sampling not only enable us to produce pose-consistent target images, but also improve their text-image correspondence by leveraging the full expressiveness of the text-to-image diffusion model through the use of extremely high return steps. This approach enables us to adapt 3D generators to domains with significant domain gaps without the need for time-consuming CLIP- and pose reconstruction-based filtering processes in DATID-3D~\cite{kim2022datid}, making the pipeline more efficient and simplified.

\subsection{Text-guided debiasing}
We observe that text-to-image diffusion models often suffer from an instance bias issue where only a few instances representing the text prompts are generated in the images. However, when we specify subclasses (e.g. breeds of dog) of the objects represented by the text prompt, the images of the instance are synthesized well. Based on this observation, we propose a text-guided debiasing method to improve intra-domain diversity, as depicted in Fig.~\ref{fig4}.

To debias the target domain $\mathcal{X}^{\text{tda}}$ represented by the text $y^{\text{tda}}$ in terms of attribute $\mathcal{A}$, we first obtain a set $\{y^{\text{tda},\mathcal{A}}_i\}_{i=1}^{N^{\text{sub}}}$ of $N^{\text{sub}}$ subclass texts $y^{\text{tda},\mathcal{A}}_i$ from various sources such as books, web search, or AI-powered chatbots like ChatGPT. 
We can ask ChatGPT "Tell me the list of \textit{Breeds} of \textit{Dogs}." when $\mathcal{X}^{\text{tda}}=\textit{Dog}$ and $\mathcal{A}=\textit{Breed}$. 
Then, we generate pose-aware target dataset $\mathcal{D}(y^{\text{tda},\mathcal{A}}_i)$ for each subclass text. 
Finally, combining these dataset, we can construct the debiased dataset $\mathcal{D}^{\text{deb}}(y^{\text{tda}},\mathcal{A}) = \{\mathcal{D}(y^{\text{tda},\mathcal{A}}_i)\}_{i=1}^{N^{\text{sub}}}$.

\begin{figure}[!t]
    \centering
    \includegraphics[width=\linewidth]{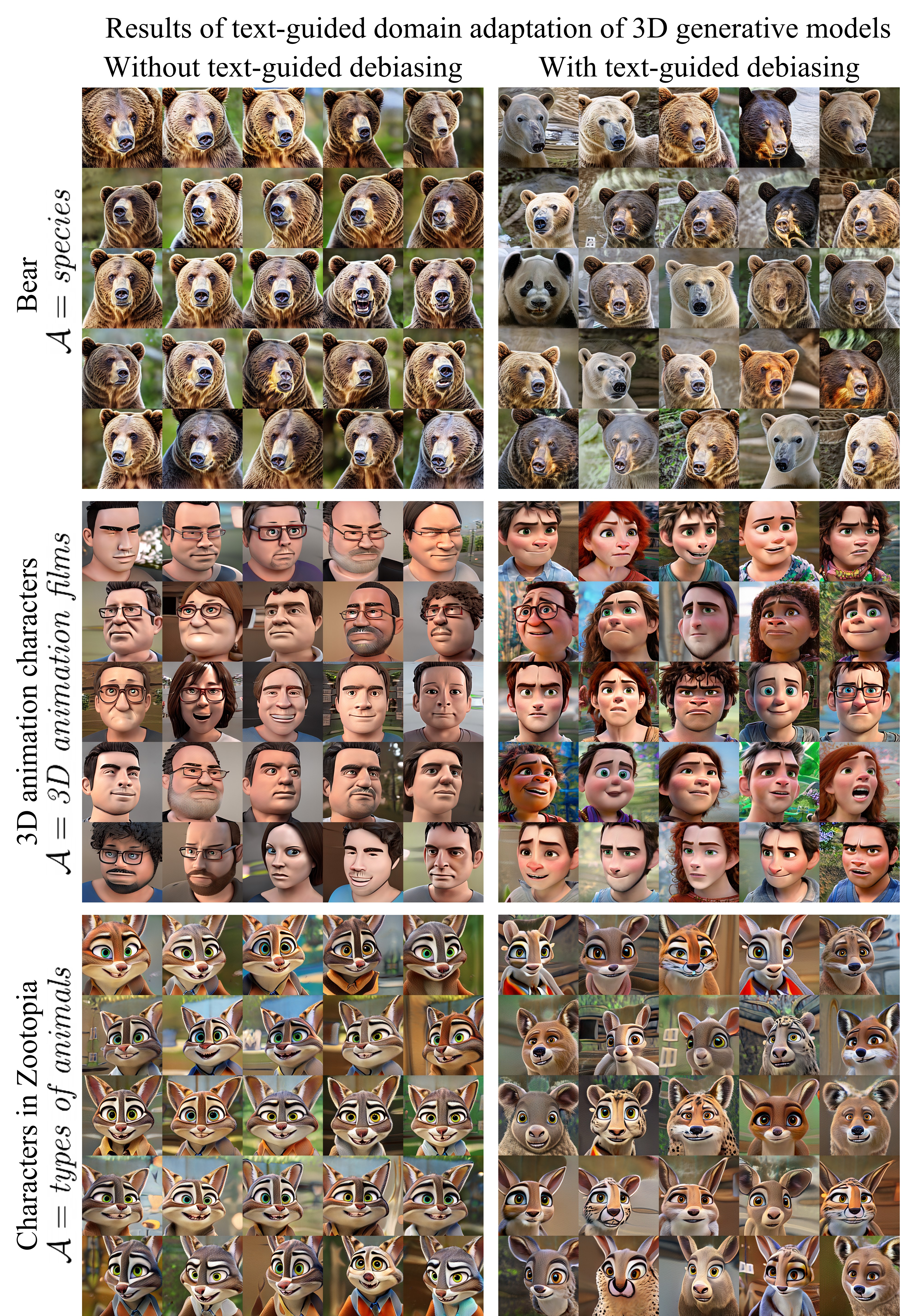}
    \vspace{-1.5em}
    \caption{Our text-guided debiasing method improves intra-domain diversity of the results of text-guided domain adaptation.}
    \vspace{-1.5em}
    \label{fig7}
\end{figure}

\begin{figure*}[!t]
    \centering
    \includegraphics[width=\linewidth]{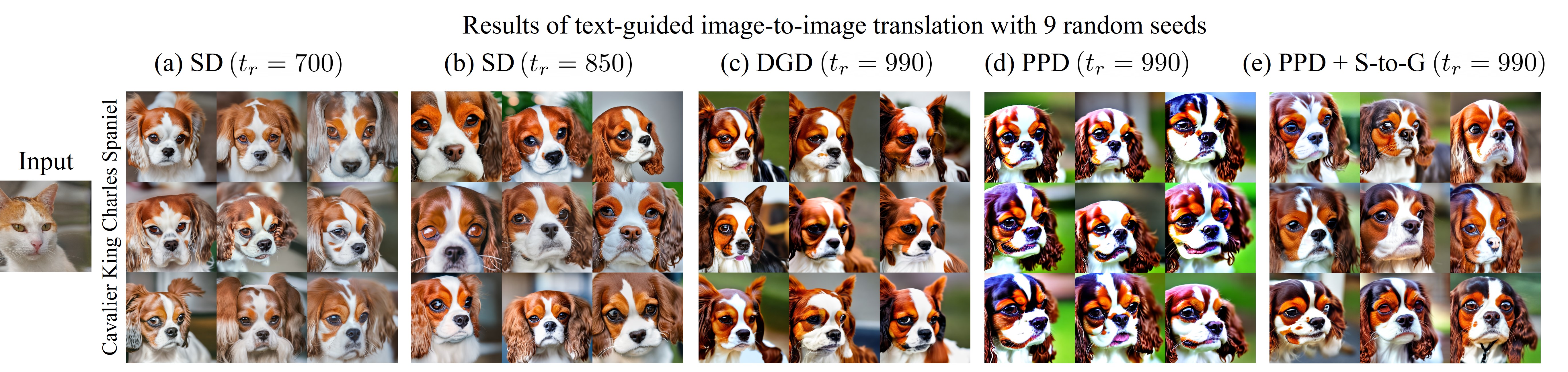}
   \vspace{-2.em}
    \caption{Results of text-guided image-to-image translation using Stable diffusion (SD)~\cite{rombach2022high}, depth-guided diffusion (DGD)~\cite{rombach2022high}, our pose-preserved diffusion (PPD), and specialized-to-general (S-to-G). Pose-preserved diffusion enable image translation with pose-consistency and domain adaptation with high-quality of 3D shapes. S-to-G allows to resolve the bias issue in details.}
   \vspace{-1em}
    \label{fig8}
\end{figure*}

\begin{figure}[!t]
    \centering
    \includegraphics[width=\linewidth]{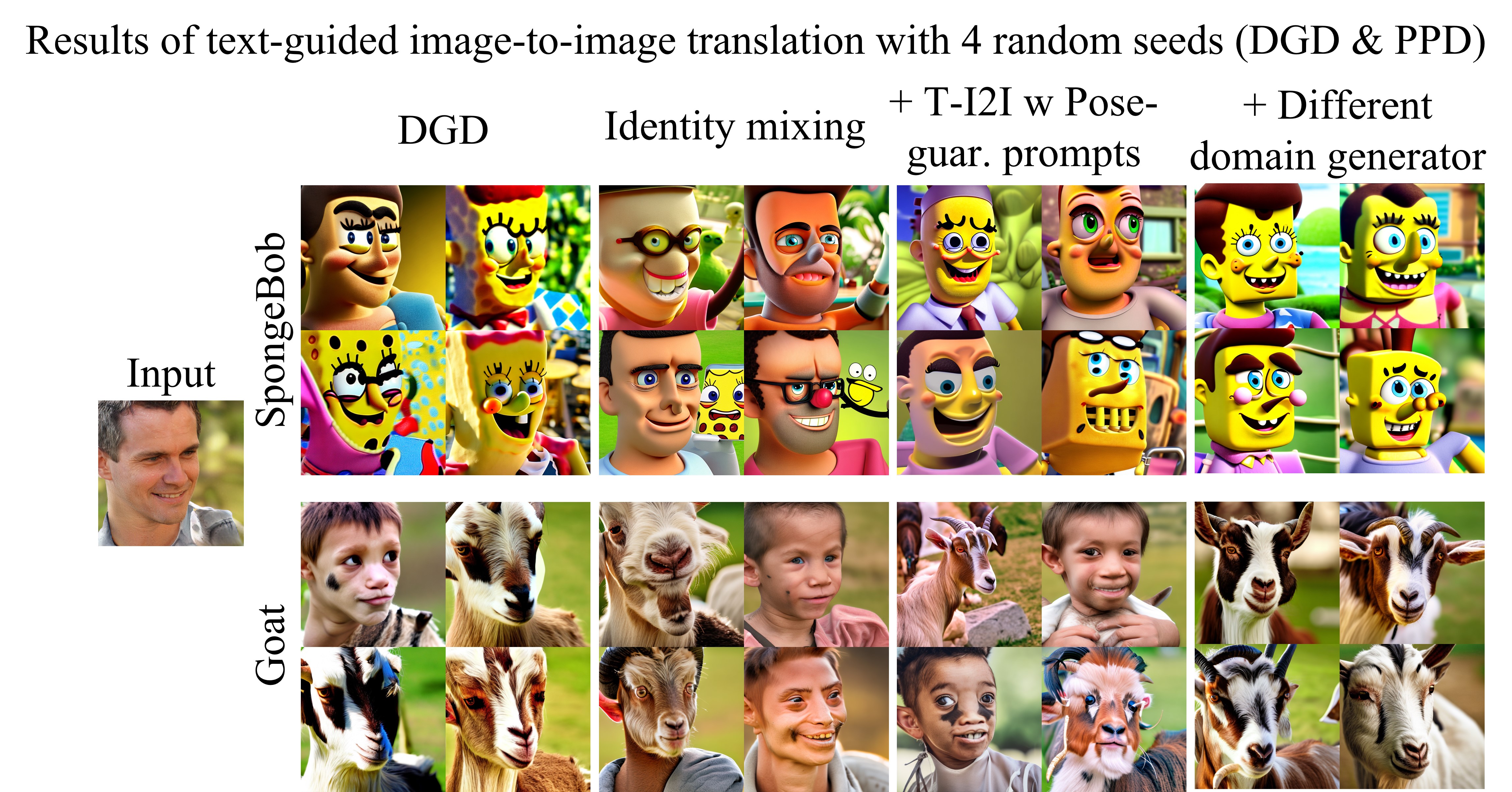}
   \vspace{-2.em}
    \caption{Results of text-guided image-to-image translation depending on the data collection stratgies for training the pose-preserved diffusion model.}
   \vspace{-1em}
    \label{fig9}
\end{figure}

\section{Experiments}
\label{sec3_experiments}
We demonstrate the effectiveness of our approach by applying it to a range of diverse domains with significant domain gaps using state-of-the-art 3D generators, EG3D~\cite{chan2022efficient}.
For the experiments, we employ a Stable diffusion and its variants, depth-guided diffusion~\cite{rombach2022high}.
We use MiDaS~\cite{Ranftl2021,Ranftl2022} as our depth map estimation model.
To fine-tune the 3D generoators, 1,000 target images per text prompt are used. 
We set $\eta=0.4$ as default.
In case of text-guided debiased dataset, we use 300 images per subclass text.
For more detailed information about the setup of experiments, see the supplementary Section~\ref{supp_implementation_details} and Section~\ref{supp_experimental_details}.

\subsection{Evaluation}

\begin{table}[!tb]
\centering
\caption{User study results on text-image correspondence, realism and diversity of rendered images from adapted generators.}\label{tab1}%
\vspace{-0.5em}
\label{tab1}
\centering
\begin{adjustbox}{width=0.8\linewidth}
\begin{tabular}{lccc}
\toprule
\textbf{Rendered 2D images}            & Text-Corr.↑    & Realism↑       & Diversity↑   \\
 \midrule
StyleGAN-NADA$^{*}$ & 3.267 & 2.571 & 2.719 \\
HyperDomainNet$^{*}$ & 3.231 & 2.576 & 2.776 \\
StyleGANFusion & 3.502 & 2.812 & 2.871 \\
DATID-3D & 3.776 & 3.148 & 3.160 \\
\textbf{Ours} & \textbf{4.071} & \textbf{3.455} & \textbf{3.426} \\
\bottomrule
\end{tabular}
\end{adjustbox}
\vspace{-0.5em}
\end{table}

\begin{table}[!tb]
\centering
\caption{User study results on text-image correspondence, sense of depth and details of 3D shape extracted from adapted generators.}\label{tab2}%
\vspace{-0.5em}
\label{tab2}
\centering
\begin{adjustbox}{width=0.8\linewidth}
\begin{tabular}{lcc}
\toprule
\textbf{3D shapes} & Text-Corr.↑    & Sense of depth \& Details↑       \\
 \midrule
StyleGAN-NADA$^{*}$ & 2.707 & 2.779 \\
HyperDomainNet$^{*}$& 2.688 & 2.802 \\
StyleGANFusion & 2.860 & 2.981 \\
DATID-3D & 3.214 & 3.260 \\
\textbf{Ours}  & \textbf{3.495}     & \textbf{3.440}   \\
\bottomrule
\end{tabular}
\end{adjustbox}
\vspace{-1em}
\end{table}

\paragraph{Baselines.}
We compare our approach to several recent methods for domain adaptation in 3D generative models, including StyleGANFusion~\cite{song2022diffusion} and DATID-3D~\cite{kim2022datid}, both of which are based on text-to-image diffusion methods. We also compare our approach to CLIP-based methods for 2D generative models, StyleGAN-NADA~\cite{radford2021learning} and HyperDomainNet~\cite{alanov2022hyperdomainnet}, denoted by a star symbol ($^*$) to indicate their extension to 3D models.

To evaluate our method, we use the EG3D~\cite{chan2022efficient} generator pretrained on $512^2$ images from the FFHQ dataset~\cite{karras2019style} as our source generator, and adapt it to a range of diverse domains with significant domain gaps. In contrast, other methods used the EG3D generator pretrained on $512^2$ FFHQ images for adaptation to movie or animation characters, and the EG3D generator pretrained on $512^2$ AFHQ-cat images for adaptation to animal domains, following their original experimental settings. 

\paragraph{Qualitative results.}
As shown in Fig.~\ref{fig1} and~\ref{fig5}, 
Our method successfully adapts 3D generators to the domains with significant domain gaps, enabling the synthesis of diverse samples with excellent text-image correspondence and 3D shapes. In contrast, other methods fail to adapt to these domains. For instance, when the target domain is an elephant, the samples and 3D shapes generated by StyleGAN-NADA$^*$~\cite{radford2021learning}, HyperDomainNet$^*$\cite{alanov2022hyperdomainnet}, and StyleGANFusion\cite{song2022diffusion} resemble cats more than elephants. Although DATID-3D succeeds in generating samples that resemble elephants, its pose is biased toward the front view, leading to poor quality of 3D shapes. In comparison, our method produces images that closely correspond to elephant images with detailed shapes. 

\paragraph{User study.}
We conduct a user study to evaluate the quality of the generated samples and 3D shapes from the shifted generator through baselines and our methods and report the mean opinion score. The participants were requested to assess the visual quality of the generated images in terms of text-image consistency, realism, and diversity using a rating scale ranging from 1 to 5. Additionally, we asked users to rate the text-image correspondence and sense of depth \& details for evaluating the 3D shapes. Our results, presented in Table~\ref{tab1} and Table~\ref{tab2}, demonstrate the superior text-image correspondence, realism, diversity, and quality of 3D shapes compared to the baselines. See the supplementary Section~\ref{supp_experimental_details} for further details on the comparison.

\subsection{Text-guided debiasing}
We apply text-guided debiasing to bear, 3D animation characters, and characters in Zootopia using the attribute species, 3D animation characters, and types of animals, respectively. We obtained the information on subclasses from ChatGPT.
As represented in Fig.~\ref{fig7}, our text-guided debiasing method enable to enhance intra-domain diversity (Bear, Characters in Zootopia) or further improve the text-image correspondence (3D animation characters).

\subsection{Ablation studies}
\label{subsec_exp_ablation}

\subsection{Pose-preserved text-to-image diffusion}
In Fig.~\ref{fig8}, we compare the results of text-guided image translation using Stable diffusion (SD), the depth-guided diffusion (DGD), our proposed pose-preserved diffusion (PPD), and our specialized-to-general sampling strategy (S-to-G). We observe that SD exhibits low image-text correspondence, such as unnatural ear shapes when using a low return step, and low pose consistency when using a high return step. DGD suffers from overly strong shape constraints from the depth map. In contrast, our PPD enables pose-consistent image generation but may exhibit biases in style or details. The S-to-G strategy resolves these biases by utilizing the general text-to-image diffusion for creating details.
See the supplementary Fig.~\ref{fig_supp4} for the extended results.

\paragraph{Data preparation for training PPD.}
The effectiveness of training PGD with only identity mixing is not optimal as shown in Fig.~\ref{fig9}. However, when PGD is trained with the translated targets from identity mixing using pose-guaranteed prompts, PPD enables large shape changes, particularly when the target domain is similar to the human domain (e.g. SpongeBob). However, PPD fails to produce satisfactory results when the domain gap is significant (e.g. Goat). To overcome this limitation, we leverage another generator trained on a different domain, which enables PPD to translate the input image to target images even with a large domain gap.

\section{Discussion and Conclusion}
\paragraph{Limitation.}
\label{subsec_exp_analysis}
Our methods may pose potential societal risks and therefore should be used with caution for appropriate purposes. Further information on limitations and potential negative social impacts can be found in the supplementary Section~\ref{supp_additional_discussion}.

\paragraph{Conclusion.}
We propose a novel pipeline called PODIA-3D, for domain adaptation of 3D generative models using pose-preserved text-to-image diffusion models. By utilizing PPD and specialized-to-general sampling models, our method is able to adapt 3D generators to the domains across large domain gaps, broadening applicability of 3D generative models. Our method achieves superior text-image correspondence and 3D shapes compared to existing methods. Additionally, we propose a text-guided debiasing method to address instance bias.

\section*{Acknowledgements}

This work was supported by the National Research Foundation of Korea(NRF) grants funded by the Korea government(MSIT) (NRF-2022R1A4A1030579, NRF-2022M3C1A309202211), Basic Science Research Program through the NRF funded by the Ministry of Education(NRF-2017R1D1A1B05035810) and a grant of the Korea Health Technology R\&D Project through the Korea Health Industry Development Institute (KHIDI), funded by the Ministry of Health \& Welfare, Republic of Korea (grant number: HI18C0316).

{\small
\bibliographystyle{ieee_fullname}
\bibliography{egbib}
}


\clearpage
\appendix

\twocolumn[{%
\renewcommand\twocolumn[1][]{#1}%

\begin{center}
\bigskip 
\bigskip 
\textbf{\Large PODIA-3D: Domain Adaptation of 3D Generative Model Across \\ Large Domain Gap Using Pose-Preserved Text-to-Image Diffusion \\ (Supplementary Material) \\}

\bigskip 
\bigskip 
{\large  Gwanghyun Kim$^1$ \qquad Ji Ha Jang$^1$ \qquad Se Young Chun$^{1,2,\dagger}$ \\
$^1$Dept. of Electrical and Computer Engineering, $^2$INMC \&  IPAI \\
Seoul National University, Republic of Korea \\
{\tt\small \{gwang.kim, jeeit17, sychun\}@snu.ac.kr}
}
\bigskip 
\bigskip 
\maketitle
 
\end{center}%
}]
{
  \renewcommand{\thefootnote}%
    {\fnsymbol{footnote}}
  \footnotetext[2]{Corresponding author.}
}

\setcounter{equation}{0}
\setcounter{figure}{0}
\setcounter{table}{0}
\setcounter{page}{1}
\makeatletter
\renewcommand{\theequation}{S\arabic{equation}}
\renewcommand{\thefigure}{S\arabic{figure}}
\renewcommand{\thetable}{S\arabic{table}}

\section{Videos}

We provide supplementary videos that presents a more comprehensive visualization and demonstration of the effectiveness of our PODIA-3D method in adapting 3D generators across significant domain gaps at \href{https://gwang-kim.github.io/podia_3d}{\small{\texttt{gwang-kim.github.io/podia\_3d}}}. The videos showcases the high level of text-image correspondence achieved as well as the high quality of 3D shapes, achieved by our approach.


\section{Details on Methods}
\label{supp_methods_details}

\subsection{SD vs DGD vs PPD}
Stable diffusion (SD)~\cite{rombach2022high}  is a latent-based text-to-image diffusion model.
This model is composed of frozen VQGAN~\cite{esser2021taming} encoder \& decoder and a noise prediction model $\epsilonb_\phi^{\text{SD}}$, which is conditioned on time and text.
The VQGAN encoder~\cite{esser2021taming} $E^V$ encodes an image $\xb$ into a latent vector $\qb_0=E^V(\xb)$ and the decoder $D^V$ converts the latent to the reconstructed image $\hat{\xb}=D^V(\qb_0)$.
To train the noise prediction model $\epsilonb_\phi$, the latent $\qb_0$ is first perturbed into $\qb_t=\sqrt{\bar{\alpha}_t}\qb_0 + \sqrt{1-\bar{\alpha}_t}\epsilonb$ through the forward diffusion~\cite{ho2020denoising}, where $\epsilonb \sim \mathcal{N}(0,1)$, $t \sim \mathcal{U}([1,T])$, $\bar{\alpha}_t$ is noise schedule, and $T$ is the total diffusion steps. In SD, $T$ is set to $1,000$.
Then, the noise prediction model $\epsilonb_\phi^{\text{SD}}$ is trained to predict the noise $\epsilonb$ included in $\qb_t$, given $\qb_t, t$ and the text prompt $y$ representing $\xb$, using following objective:
\begin{align}
\mathbb{E}_{\xb,  y, \epsilonb, t }[\|\epsilonb-\epsilonb_\phi^{\text{SD}}(\qb_t, y, t)\|_2^2], \nonumber
\end{align}
where $(\xb,  y)$ are image-text pairs. 
In the noise prediction model $\epsilonb_\phi^{\text{SD}}$, the text prompt $y$ is encoded to the text embedding through the CLIP~\cite{radford2021learning} text encoder and the time $t$ is converted to the time feature using the Fourier feature~\cite{tancik2020fourfeat}.
To enable classifier-free guidance~\cite{ho2022classifier}, a single diffusion model is trained on conditional and unconditional objectives by randomly dropping $y$ to  $\varnothing$.
The noisy latent $\qb_t$ is given to the input of the first convolution layer of the UNet-based~\cite{ronneberger2015u} autoencoder while the time feature and the text embedding are conditioned on the normalization layers through cross-attention~\cite{vaswani2017attention} mechanism.

\begin{figure}[!tb]
    \centering
    \vspace{-0em}
    \includegraphics[width=0.96\linewidth]{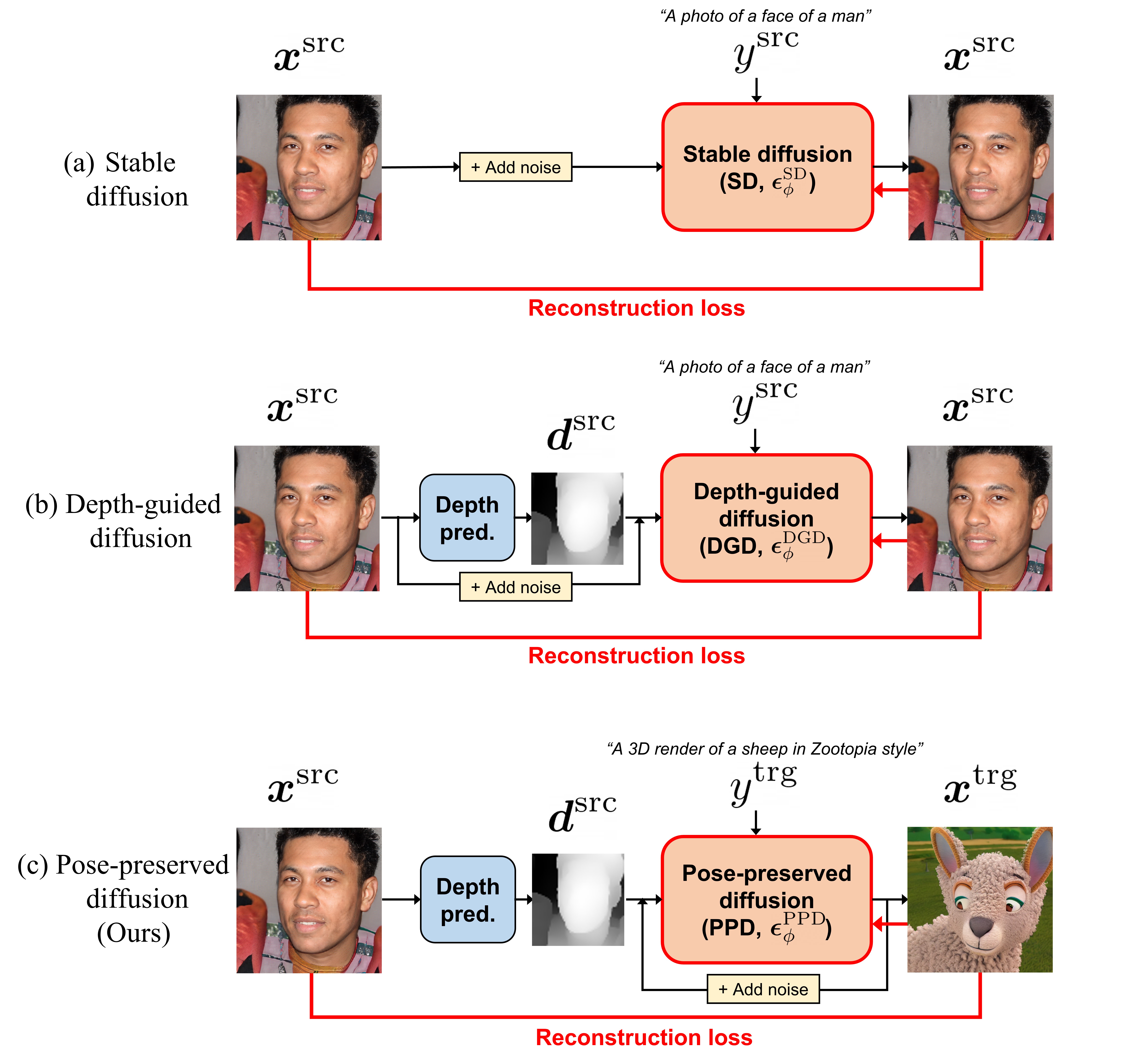}
    \caption{Comparison of training process for 3 different diffusion models: (a) Stable diffusion~\cite{rombach2022high}, (b) depth-guided diffusion~\cite{rombach2022high} and (c) our pose-preserved diffusion.}
    \label{fig_supp1}
\end{figure}

Depth-guided diffusion model (DGD)~\cite{rombach2022high} is a variant of SD~\cite{rombach2022high}, where the depth map $\db=h(\xb)$ is predicted from $\xb$ using the pretrained depth estimation model $h$ and concatenated with $\qb_t$, and used as input to the UNet~\cite{ronneberger2015u} autoencoder as an extra conditioning. 
The DGD noise prediction model $\epsilonb_\phi^{\text{DGD}}$ is fine-tuned from the pretrained $\epsilonb_\phi^{\text{SD}}$ using following objective: 
\begin{align}
\mathbb{E}_{\xb,  y, \epsilonb, t }[\|\epsilonb-\epsilonb^{\text{DGD}}_\phi(\qb_t, y, \db, t)\|_2^2], \nonumber
\end{align}
where $(\xb,  y)$ are image-text pairs.  
DGD is fine-tuned from the pretrained SD.

\setlength{\textfloatsep}{1em}%

\begin{algorithm}[!t]
    \caption{Text-guided image-to-image translation (T-I2I)\label{algo1}}
    \DontPrintSemicolon
    \SetAlgoNoLine
    \SetAlgoVlined
    \SetKwProg{Fn}{Require}{:}{}
    \SetKwProg{Fn}{Function}{:}{}
    \SetKwFunction{TextGuidedItoI}{T\_I2I}
    \KwIn{$\epsilonb_{\phi} \in \{\epsilonb^{\textnormal{SD}}_{\phi}, \epsilonb^{\textnormal{DGD}}_{\phi}, \epsilonb^{\textnormal{PPD}}_{\phi}\}$, $E^{V}$, $ D^{V}$, $y$, $t_r$, $s$, *}
    \Fn{\TextGuidedItoI{$\xb^{\textnormal{src}}$, $y$, $\epsilonb_{\phi}$, *}}
    {
    
    $\qb_{0}=E^{V}(\xb^{\textnormal{src}})$,  $\epsilonb \sim \mathcal{N} (\mathbf{0,I})$ \;
    $\qb^{\textnormal{trg}}_{t_r} = \sqrt{\bar\alpha_{t_r}}\qb_{0}  + \sqrt{1 - \bar\alpha_{t_r}}\epsilonb$ \;
    \If{$\epsilonb_{\phi} \in \{\epsilonb^{\textnormal{DGD}}_{\phi}, \epsilonb^{\textnormal{PPD}}_{\phi}\}$}{
    $\db=h(\xb^{\textnormal{src}})$ \;
    }
    \For{$\ t = t_r, t_r-1,\ldots, 1$}{
        \small
        \uIf{$\epsilonb_{\phi} \in \{\epsilonb^{\textnormal{DGD}}_{\phi}, \epsilonb^{\textnormal{PPD}}_{\phi}\}$}{
    $\tilde{\epsilonb} = s\epsilonb_\phi(\qb^{\textnormal{trg}}_{t},y,\db,{t}) \ \setminus \newline +  (1-s)\epsilonb_\phi(\qb^{\textnormal{trg}}_{t},\varnothing,\db,{t})$\;
    }
    \Else{
        $\tilde{\epsilonb} = s\epsilonb_\phi(\qb^{\textnormal{trg}}_{t},y,{t}) + (1-s)\epsilonb_\phi(\qb^{\textnormal{trg}}_{t},\varnothing,{t})$\;
        }
        $\qb^{\textnormal{trg}}_{t-1} = \textnormal{Sampling}(\qb^{\textnormal{trg}}_{t}, \tilde{\epsilonb}, t)$ \;
        }
   \normalsize
    $\xb^{\textnormal{trg}}=D^{V}(\qb^{\textnormal{trg}}_{0})$ \;
    \KwRet{$\xb^{\textnormal{trg}}$}
    } 

\end{algorithm}

Our pose-preserving diffusion model (PPD) has the same architecture as DGD, but it is trained differently.
It uses the depth map $\db^{\text{src}}=h(\xb^{\text{src}})$ from the source image $\xb^{\text{src}}$, but the latent $\qb_0^{\text{trg}}=E^V(\xb^{\text{trg}})$ from the target image $\xb^{\text{trg}}$, giving the noisy latent $\qb_t^{\text{trg}}=\sqrt{\bar{\alpha}_t}\qb_0^{\text{trg}} + \sqrt{1-\bar{\alpha}_t}\epsilonb$.
The PPD noise prediction model $\epsilonb_\phi^{\text{PPD}}$ is fine-tuned from the pretrained $\epsilonb_\phi^{\text{DGD}}$ using following objective: 
\begin{align}
\mathbb{E}_{\xb^{\text{src}}, \xb^{\text{trg}}, y^{\text{trg}}, \epsilonb, t}[\|\epsilonb-\epsilonb^{\text{PPD}}_\phi(\qb^{\text{trg}}_t, y^{\text{trg}}, \db^{\text{src}}, t)\|_2^2], \nonumber
\end{align}
where $(\xb^{\text{src}}, \xb^{\text{trg}}, y_{\text{trg}})$ are a set of the source image, the target image and the target text.

Text-guided image-to-image translation (T-I2I) is a technique that combines diffusion-based image-to-image translation~\cite{meng2021sdedit} with text-to-image diffusion models~\cite{rombach2022high, ramesh2022hierarchical, saharia2022photorealistic}. 
The generalized T-I2I algorithm for $\epsilonb^{\text{SD}}_{\phi}, \epsilonb^{\text{DGD}}_{\phi}$ or $ \epsilonb^{\text{PPD}}_{\phi}$ is described in Algorithm~\ref{algo1}.
Initially, we convert $\xb^{\text{src}}$ to $\qb_0=E^{V}(\xb^{\text{src}})$ using the VQGAN encoder~\cite{esser2021taming} $E^V$ and perturb it to generate $\qb^{\text{trg}}_{t_r}$ where $t_r \in [1,T]$ is the return step.
Next, $\qb_{0}^{trg}$ is obtained from the noisy latent $\qb_{t_r}^{\text{trg}}$ by proceeding the sampling process  using $\epsilonb_{\phi} \in \{\epsilonb^{\text{SD}}_{\phi}, \epsilonb^{\text{DGD}}_{\phi}, \epsilonb^{\text{PPD}}_{\phi}\}$.
The guidance scale $s$ is used to adjust the scale of gradients from the target prompt $y$ and the empty prompt $\varnothing$.
Finally, we can obtain the target image $\xb^{\text{trg}}=D^{V}(\qb_{0}^{trg})$ using the VQGAN decoder~\cite{esser2021taming} $D^V$.
Any sampling method such as DDPM~\cite{ho2020denoising}, DDIM~\cite{song2020denoising}, PLMS~\cite{liu2022pseudo} can be used. 

The comparison of the training process for each model is shown in Fig.~\ref{fig_supp1}.

\begin{algorithm}[!t]
    \caption{Specialized-to-general sampling\label{algo2}}
    \DontPrintSemicolon
    \SetAlgoNoLine
    \SetAlgoVlined
    \SetKwProg{Fn}{Require}{:}{}
    \SetKwProg{Fn}{Function}{:}{}
    \SetKwFunction{StoG}{S\_to\_G}
    \KwIn{$\epsilonb^{\textnormal{SD}}_{\phi}, \epsilonb^{\textnormal{PPD}}_{\phi'}$, $E^{V}$, $ D^{V}$, $y$, $t_r$, $\eta$, $s$,$h$ *}
    \Fn{\StoG{$\xb^{\textnormal{src}}$, $y$, $\epsilonb^{\textnormal{SD}}_{\phi}, \epsilonb^{\textnormal{PPD}}_{\phi'}$, *}}
    {
    
    $\qb_{0}=E^{V}(\xb^{\textnormal{src}})$,  $\epsilonb \sim \mathcal{N} (\mathbf{0,I})$ \;
    $\qb^{\textnormal{trg}}_{t_r} = \sqrt{\bar\alpha_{t_r}}\qb_{0}  + \sqrt{1 - \bar\alpha_{t_r}}\epsilonb$ \;
    $\db=h(\xb^{\textnormal{src}})$\;
    \For{$\ t = t_r, t_r-1,\ldots, 1$}{
    \small
        \uIf{$t_r > (1-\eta)T$}{
    $\tilde{\epsilonb} = s\epsilonb^{\textnormal{PPD}}_{\phi'}(\qb^{\textnormal{trg}}_{t},y,\db,{t}) \ \setminus \newline  + (1-s)\epsilonb^{\textnormal{PPD}}_{\phi'}(\qb^{\textnormal{trg}}_{t},\varnothing,\db,{t})$\;
    }
    \Else{
        $\tilde{\epsilonb} = s\epsilonb^{\textnormal{SD}}_\phi(\qb^{\textnormal{trg}}_{t},y,{t}) + (1-s)\epsilonb^{\textnormal{SD}}_\phi(\qb^{\textnormal{trg}}_{t},\varnothing,{t})$\;
        }
        
        $\qb^{\textnormal{trg}}_{t-1} = \textnormal{Sampling}(\qb^{\textnormal{trg}}_{t}, \tilde{\epsilonb}, t)$ \;
        }
    \normalsize
    $\xb^{\textnormal{trg}}=D^{V}(\qb^{\textnormal{trg}}_{0})$ \;
    \KwRet{$\xb^{\textnormal{trg}}$}
    } 
\end{algorithm}

\subsection{Specialized-to-general sampling}
In the initial $\eta T$ phase, with $\eta \in [0,1]$ representing the PPD ratio, we leverage the PPD model to produce major structural components and pose information. 
For the remaining $(1-\eta)T$ duration, we employ Stable diffusion models to generate smaller image details or structures. 
The specialized-to-general sampling process is outlined in Algorithm~\ref{algo2}.

\subsection{Text-guided adaptation of 3D generative models}
We utilize the total loss $\mathcal{L}_{\text{total}} = \mathcal{L}_{\text{ADA}} + \lambda \mathcal{L}_{\text{den}}$, which includes both the ADA loss and the density regularization loss.
Here are the details on the losses used for adversarially fine-tuning the 3D generator following the fine-tuning process in~\cite{kim2022datid, chan2022efficient}.
\paragraph{ADA loss.}  
The ADA loss, $\mathcal{L}_{\text{ADA}}$~\cite{karras2020training}, is an adversarial loss that incorporates adaptive dual augmentation and R1 regularization. Specifically, the loss is defined as follows:
\small
\begin{align}
    &\mathcal{L}_{\text{ADA}}=\mathbb{E}_{\zb \sim \mathcal{Z}, \cb \sim \mathcal{C}} [f(D_\omega(A(G_\theta(\zb, \cb)), \cb)]  \\
    &+\mathbb{E}_{(\cb, \xb^{\text{tda}}) \in \mathcal{D}^{\text{tda}}}[f(-D_{\omega}(A(\xb^{\text{tda}}), \cb)+\lambda\|\nabla D_{\omega}(A(\xb^{\text{tda}}), \cb)\|^2)] \nonumber
\end{align}
\normalsize
where $A$ is a stochastic non-leaking augmentation operator with probability $p$ and $f(u)=-\log (1+\exp (-u))$.

\paragraph{Density regularization loss.}
We also employ density regularization, which has proven to be useful in mitigating the occurrence of undesired shape distortions by encouraging smoothness in the density field~\cite{chan2022efficient}.
For each rendered scene, we randomly select points $v$ from the volume $\mathcal{V}$, as well as additional perturbed points that have been slightly distorted by Gaussian noise $\delta v$. We then compute the L1 loss between the predicted densities, as shown below:
\begin{align}
    \mathcal{L}&_{\text{den}}= \mathbb{E}_{v\in\mathcal{V}}[\|\sigma_\theta(v) - \sigma_\theta(v+\delta v)\|].
\end{align}

Algorithm for our text-guided adaptation of 3D generative models is provided in Algorithm~\ref{algo3}.
 We translate the source image $\xb^{\text{src}}$ to yield the target image $\xb^{\text{tda}}$ guided by a text prompt $y$ using PPD and specialized-to-general sampling, constructing a set of $(\cb, \xb^{\text{tda}})$. 
Initially, we duplicate the pre-trained 3D generator, $G_\theta$, to create $G_{\theta'}$. We also initialize a pose-conditioned discriminator, $D_\omega$.
For $i=1,2,...,N$, we start by sampling a random latent vector $\zb$ and camera parameter $\cb$. We then compute the ADA loss for the generator, denoted as $\mathcal{L}^{G}{\text{ADA}}$, using the generator $G_{\theta'}(\zb_i, \cb_i)$, the discriminator $D_\omega$, and the stochastic non-leaking augmentation $A$. 
Additionally, for each rendered scene, we randomly select points $v$ from the volume $\mathcal{V}$ to calculate the density regularization loss $\mathcal{L}_{\text{den}}$, along with another loss. These losses are then used to update the generator.
Subsequently, we calculate the ADA losses for the discriminator using both generated images, denoted as $\mathcal{L}^{D,\textnormal{fake}}_{\text{ADA}}$, and real target images, denoted as $\mathcal{L}^{D,\textnormal{real}}_{\text{ADA}}$. These two losses are then combined to update the discriminator. We repeat this process for $K$ epochs.

\begin{algorithm}[t!]
    \caption{Text-guided adaptation of 3D generative models\label{algo3}}
    \DontPrintSemicolon
    \SetAlgoNoLine
    \SetAlgoVlined
    \SetKwProg{Fn}{Require}{:}{}
    \SetKwProg{Fn}{Function}{:}{}
    \SetKwFunction{TextGuidedItoI}{T\_I2I}
    \SetKwFunction{StoG}{S\_to\_G}
    \KwIn{$G_\theta$,  $\mathcal{D}^{\textnormal{src}}$, $\epsilonb^{\textnormal{SD}}_{\phi}$, $\epsilonb^{\textnormal{PPD}}_{\phi'}$,$K$, $A$, $f$, *}
    \KwOut{$G_{\theta'}$}

    \tcp{Pose-aware target dataset generation}
    $\mathcal{D}^{\textnormal{tda}} = \{\}$ \;
    \For{$i = 1,2,\ldots, N$}{ 
        $(\cb_i, \xb^{\textnormal{src}}_i) \in \mathcal{D}^{\textnormal{src}}$ \;
        $\xb_i^{\textnormal{tda}}$=\StoG{$\xb_i^{\textnormal{src}}$, $y$, $\epsilonb^{\textnormal{SD}}_{\phi}, \epsilonb^{\textnormal{PPD}}_{\phi'}$, *} \;
        Append $( \cb_i,\xb_i^{\textnormal{tda}})$ to $\mathcal{D}^{\textnormal{tda}}$
    }
    \BlankLine
    \tcp{Adversarial fine-tuning of 3D generators}
    $G_{\theta'} \leftarrow \textnormal{clone}(G_{\theta})$, 
    $D_\omega \leftarrow \textnormal{Initialize }D$ \;
    \For{$k = 1,2,\ldots, K$}{ 
        \For{$i = 1,2,\ldots, N$}{ 
            $\zb_i \in \mathcal{Z}$, $\cb_i  \in \mathcal{C}$, $v_i \in \mathcal{V}$ \;
            
            \tcp{Update $G_{\theta'}$}
            $\mathcal{L}^{G}_{\text{ADA}}=-f(D_\omega(A(G_{\theta'}(\zb_i, \cb_i)), \cb_i)$ \;
            
            $\mathcal{L}_{\text{den}}=\|\sigma_{\theta'}(v_i) - \sigma_{\theta'}(v_i+\delta v_i)\|$ \;
            
            $\theta' \leftarrow \textnormal{Update\_}G(\theta',\mathcal{L}^{G}_{\textnormal{ADA}} + \lambda_{\textnormal{den}}\mathcal{L}_{\text{den}})$
            
            \tcp{Update $D_{\omega}$}
            $\mathcal{L}^{D,\textnormal{fake}}_{\textnormal{ADA}}=f(D_\omega(A(G_{\theta'}(\zb_i, \cb_i)), \cb_i)$
            
            $(\cb_i, \xb_i^{\textnormal{tda}}) \in \mathcal{D}^{\textnormal{tda}} $ \;
            $\mathcal{L}^{D,\textnormal{real}}_{\textnormal{ADA}}= f(-D_{\omega}(A(\xb^{\textnormal{tda}}), \cb_i) $\\
            \quad\quad\quad\quad\quad$+\lambda\|\nabla D_{\omega}(A(\xb^{\textnormal{tda}}), \cb_i)\|^2)$
            
            $\omega \leftarrow
            \textnormal{Update\_}D(\omega,\mathcal{L}^{D,\textnormal{fake}}_{\textnormal{ADA}} + \mathcal{L}^{D,\textnormal{real}}_{\textnormal{ADA}})$

        }
    }

\end{algorithm}

\begin{figure*}[!htb]
    \centering
    \vspace{-0em}
    \includegraphics[width=0.96\linewidth]{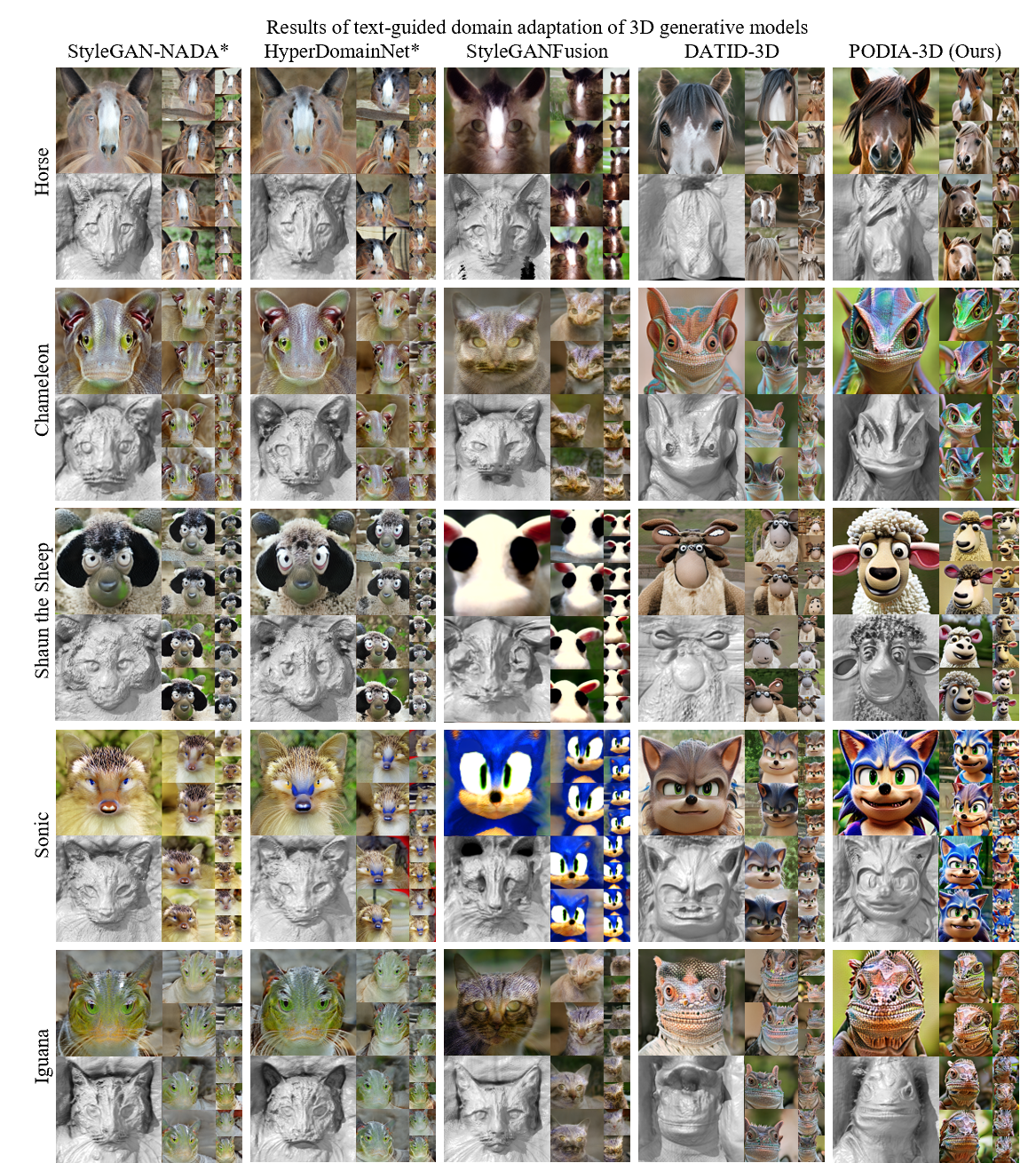}
    \caption{Qualitative comparison between our text-guided domain adaptation results and other baselines. We enable EG3D~\cite{chan2022efficient} to synthesize the multi-view consistent images in wide range of text-guided domains (animals \& characters) with high text-image correspondence and high quality of 3D shape. For the results of pose-controlled synthesis, please see the supplementary videos at \href{https://gwang-kim.github.io/podia_3d}{\small{\texttt{gwang-kim.github.io/podia\_3d}}}.}
    \label{fig_supp2}
\end{figure*}

\begin{figure*}[!htb]
    \centering
    \vspace{-0em}
    \includegraphics[width=0.96\linewidth]{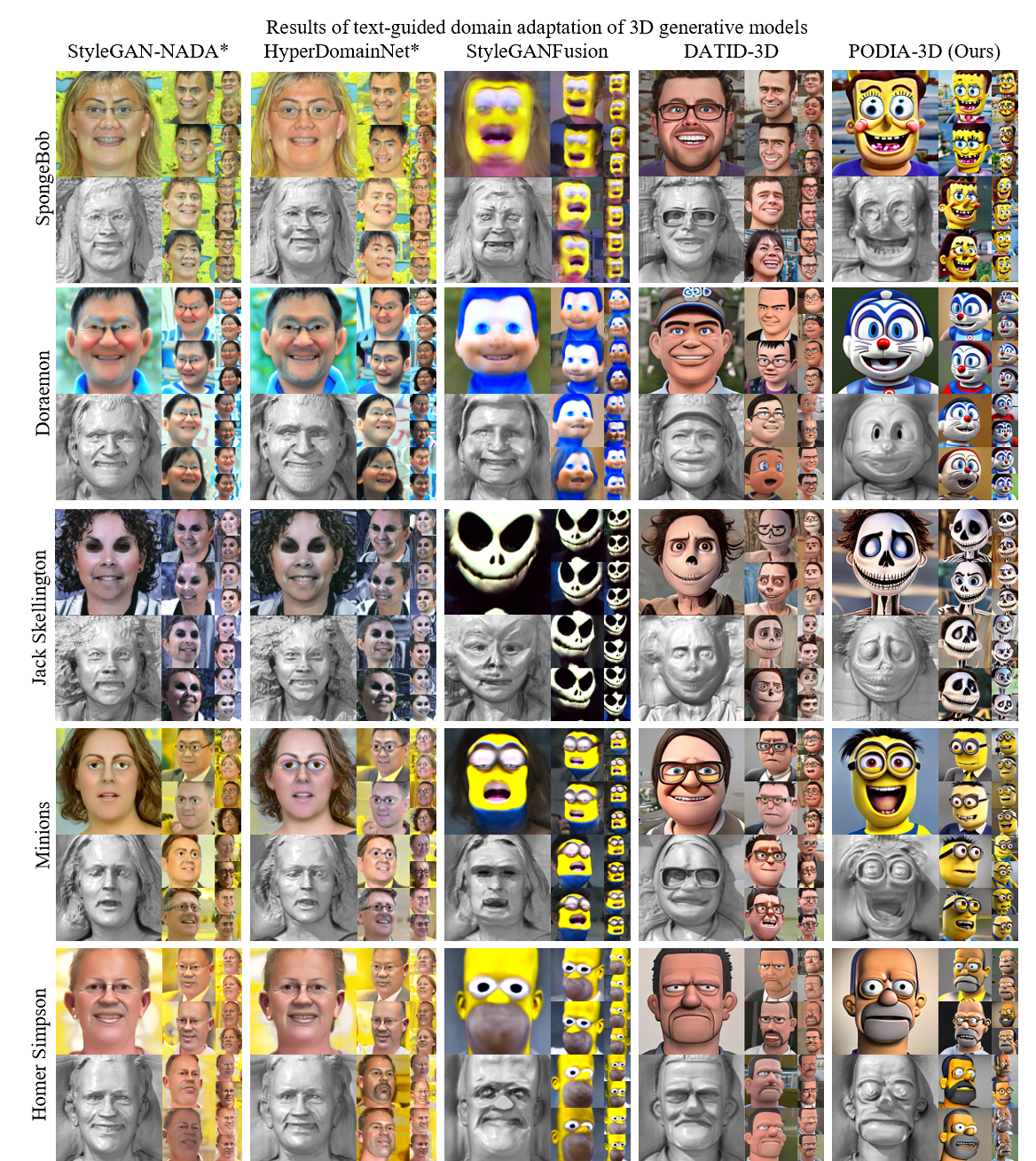}
    \caption{
    Qualitative comparison between our text-guided domain adaptation results and other baselines. We enable EG3D~\cite{chan2022efficient} to synthesize the multi-view consistent images in wide range of text-guided domains (characters) with high text-image correspondence and high quality of 3D shape. For the results of pose-controlled synthesis, please see the supplementary videos at \href{https://gwang-kim.github.io/podia_3d}{\small{\texttt{gwang-kim.github.io/podia\_3d}}}.}
    \label{fig_supp3}
\end{figure*}

\begin{figure*}[!htb]
    \centering
    \vspace{-0em}
    \includegraphics[width=0.96\linewidth]{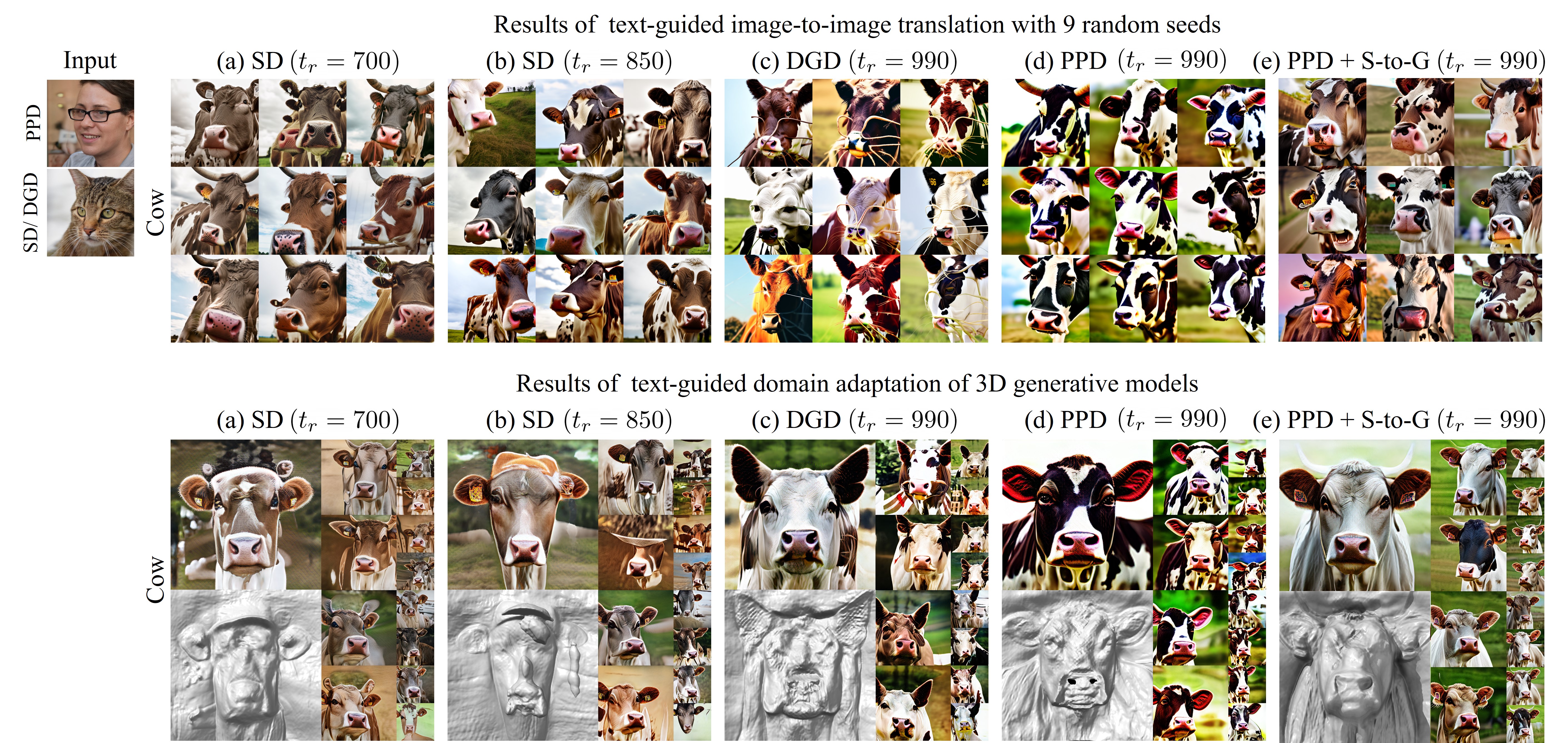}
    \caption{Results of text-guided image-to-image translation and text-guided domain adaptation of  EG3D~\cite{chan2022efficient} using Stable diffusion (SD)~\cite{rombach2022high}, depth-guided diffusion (DGD)~\cite{rombach2022high}, our pose-preserved diffusion (PPD), and specialized-to-general (S-to-G). Pose-preserved diffusion enable image translation with pose-consistency and domain adaptation with high-quality of 3D shapes. S-to-G allows to resolve the bias issue in details. }
    \label{fig_supp4}
\end{figure*}

\begin{figure}[!htb]
    \centering
    \vspace{-0em}
    \includegraphics[width=\linewidth]{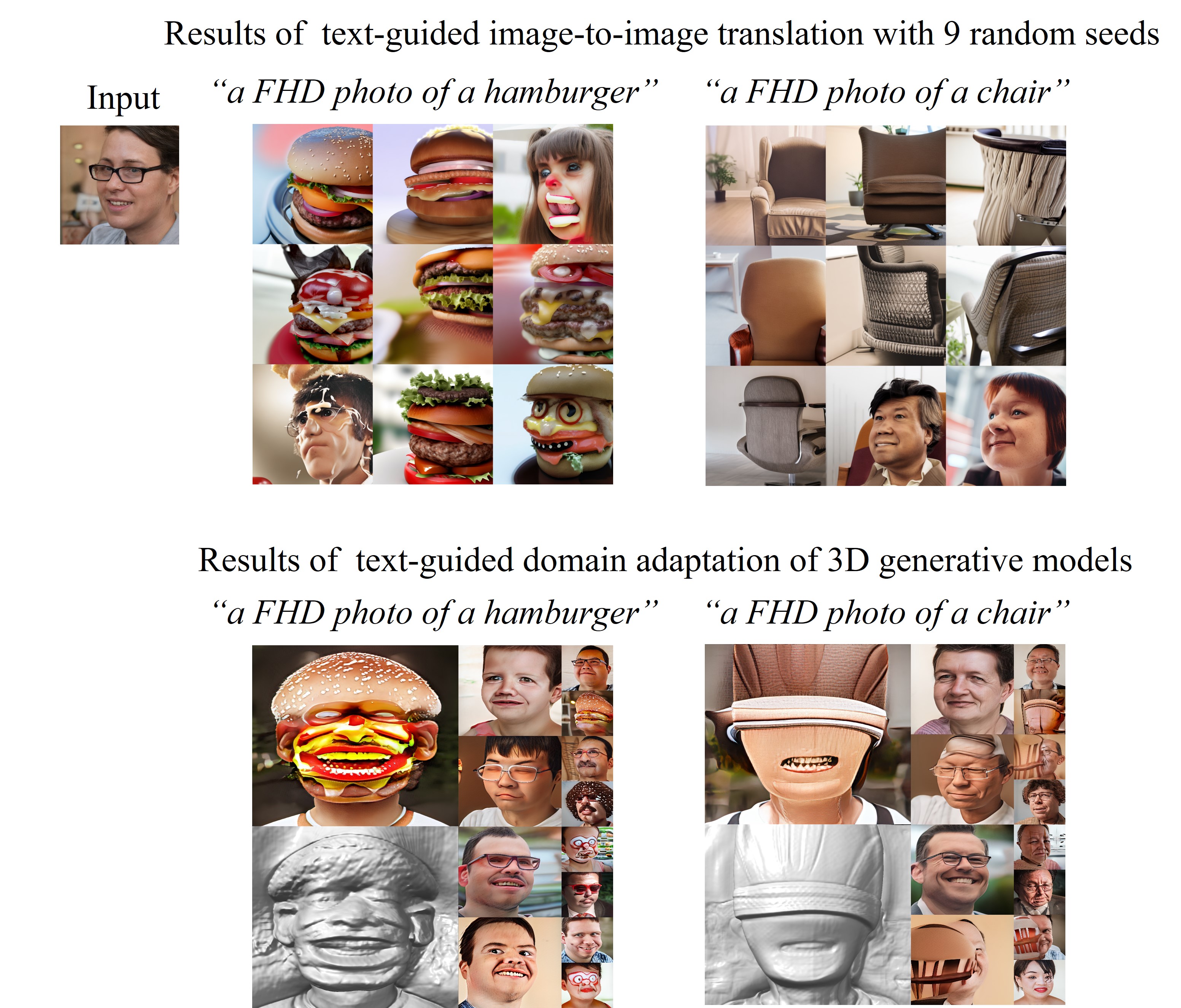}
    \caption{Domain adaptation to certain non-living objects such as chairs and hamburgers, which lack directional information, results in low text-image correspondence. }
    \label{fig_supp5}
\end{figure}

\section{Implementation Details}
\label{supp_implementation_details}
\subsection{3D generative model}
We select EG3D~\cite{chan2022efficient}, which is a stat-of-the-art 3D-aware generative model, as our source generator.
Especially, we adopt EG3D~\cite{chan2022efficient} pretrained on $512^2$ images in FFHQ~\cite{karras2019style}.
The EG3D generator is comprised of four main components: a backbone, a decoder, a volume rendering module, and a super-resolution component. The backbone features the StyleGAN2~\cite{karras2020analyzing} generator and a mapping network. The decoder, on the other hand, is a multilayer perceptron that has a single hidden layer. The super-resolution module, which utilizes two StyleGAN2 blocks, is also integrated into the generator. Lastly, the EG3D discriminator is based on a StyleGAN2 discriminator that features dual discrimination and camera pose-conditioning.

\subsection{Text-to-image diffusion models}
We utilized Stable diffusion v2~\cite{rombach2022high}, which comprises of several components such as the VQGAN encoder and decoder~\cite{esser2021taming}, UNet-based autoencoder~\cite{ronneberger2015u}, and OpenCLIP-ViT/H text encoder~\cite{radford2021learning,schuhmann2022laionb, ilharco_gabriel_2021_5143773}. To train this model, we used a filtered subset of LAION-5B~\cite{schuhmann2022laionb}. Specifically, the model underwent training for 550k steps on $256^2$ images, 1,000k steps on $512^2$ images, and 140k steps on $768^2$ images.

We utilize the depth-guided diffusion model~\cite{rombach2022high} for our experiments. It is trained for 550k steps on $256^2$ images and for 850k steps on $512^2$ images in a filtered subset of LAION-5B~\cite{schuhmann2022laionb}, without the use of a depth map. To incorporate depth information, we modify the UNet autoencoder~\cite{ronneberger2015u} by adding an additional input channel to receive the depth map generated by MiDaS~\cite{Ranftl2021,Ranftl2022}. The depth-guided diffusion model is fine-tuned for 200k steps on $512^2$ images in the same dataset with the updated architecture.

We utilize DEIS~\cite{zhang2022fast} as our diffusion sampling method, which is a state-of-the-art method that accelerates the diffusion process while maintaining high quality.
We configure the number of inference steps to 30 and set the return step $t_r$ and guidance scale $s$ to 990 and 10, respectively.

\subsection{Depth estimation network}
We utilize MiDaS~\cite{Ranftl2021,Ranftl2022} as our depth map estimation model, which computes relative inverse depth from a single image. This transformer-based model is trained on 12 distinct datasets, including WSVD~\cite{wang2019web}, TartanAir~\cite{tartanair2020iros}, IRS~\cite{wang2019irs}, KITTI~\cite{Geiger2013IJRR}, MegaDepth~\cite{MegaDepthLi18}, ApolloScape~\cite{huang2019apolloscape}, HRWSI~\cite{Xian_2020_CVPR}, ReDWeb~\cite{Xian_2018_CVPR}, DIML~\cite{kim2018deep}, Movies, BlendedMVS~\cite{yao2020blendedmvs}, and NYU Depth V2~\cite{Silberman:ECCV12}. The model is trained using multi-objective optimization to ensure high quality on a wide range of inputs.

\subsection{Details on training of PPD}
We generated 3,000 source images and corresponding depth maps using the pre-trained EG3D~\cite{chan2022efficient} model on FFHQ~\cite{karras2019style} and MiDaS~\cite{Ranftl2021,Ranftl2022}. 
For the target images, we first created 3,000 identity-mixed images and then generated an additional $3,000\times5$ images using text-guided image-to-image translation with 9 text prompts that guarantee pose preservation. 
To further increase the variety of target images, we collected 3,000 images using the EG3D model pre-trained on AFHQ-cat~\cite{karras2021alias, choi2020stargan} and translated them using 6 prompts, resulting in a total of $3,000\times3$ target images. 
Table~\ref{tab1} lists the text prompts used to collect the target images for training our pose-preserved diffusion model.

We conduct fine-tuning of the depth-guided diffusion models on this dataset for 2 epochs using Adam optimizer~\cite{kingma2014adam} and a batch size of 2 until the models have been trained on 50k$\sim$100k images. 

\subsection{Details on text-guided adaptation of 3D generative model}
We fine-tune the 3D generative models by using a batch size of 20 and training them until they have been exposed to 50k$\sim$100k images. Both the generator and discriminator are optimized using Adam~\cite{kingma2014adam} with a learning rate of 0.002. During training, we apply image blurring with progressively diminishing degree as the input to the discriminator, as suggested by~\cite{karras2021alias, chan2022efficient}, and we do not use style mixing. We employ ADA loss combined with R1 regularization ($\lambda=5$) and add a density regularization term with strength $\lambda_{\text{den}}=0.25$.

 Using EG3D and PPD with a DEIS~\cite{zhang2022fast} scheduler and 30 inference steps.
 We generated 1,000 target images for each text prompt in most experiments, taking an average of 30 minutes for text-guided target dataset generation. 
 We update the 3D generator with a batch size of 20 using 50k$\sim$100k training images depending on the target text prompt.

\subsection{3D shape visualization}
We utilize marching cubes to obtain iso-surfaces from the density field using marching cubes, as described in \cite{chan2022efficient}. After that, We use UCSF Chimerax~\cite{chimerax} to visualize the 3D surfaces.

\begin{table*}[!htb]
\centering
\caption{List of text prompts used to collect target images for training our pose-preserved diffusion model.}\label{tab1}%
\begin{adjustbox}{width=0.8\linewidth}
\begin{tabular}{ccc}
\toprule
Strategies                                               & Full text prompts                                                      &  \\
\midrule
Identity mixing                                          & \textit{"a FHD photo of a human face"}                                 &  \\
\midrule
\multirow{8}{*}{Pose-guranteed prompts}                  & \textit{"a 3D render of a face in Pixar style"}                        &  \\
                                                         & \textit{"a 3D render of a face of Lizardman monster in fantasy movie"} &  \\
                                                         & \textit{"a 3D render of a face of Minotaurus in fantasy movie"}        &  \\
                                                         & \textit{"a 3D render of a head of a Lego man 3D model"}                &  \\
                                                         & \textit{"a 3D render of a Stone Golem head in fantasy movie"}          &  \\
                                                         & \textit{"a FHD photo of a face of a Skeleton in fantasy movie"}        &  \\
                                                         & \textit{"a FHD photo of a face of Monkey"}                             &  \\
                                                         & \textit{"a FHD photo of white Greek statue face"}                      &  \\
\midrule
Different domain generator                               & \textit{"a FHD photo of a Cat face"}                                   &  \\
\midrule
\multirow{4}{*}{Different domain pose-guranteed prompts} & \textit{"a 3D render of a face of a Cat in Zootopia style"}            &  \\
                                                         & \textit{"a 3D render of a face of a Pig in Zootopia style"}            &  \\
                                                         & \textit{"a 3D render of a face of a Sheep in Zootopia style"}          &  \\
                                                         & \textit{"a FHD photo of a face of a Wol"}                              &  \\
\bottomrule
\end{tabular}
\end{adjustbox}
\end{table*}

\section{Experimental Details}
\label{supp_experimental_details} 
\subsection{Text prompts}
Throughout the main paper and its supplementary materials, we employ a brief text prompt to denote each individual text prompt.
The complete text prompts that corresponds to each concise prompt are summarized in Table~\ref{tab2}.
Also, the complete text prompts that correspond to each subclass prompt used for text-guided debiasing can be found in Table~\ref{tab3}.

\subsection{Baselines}
In StyleGAN-NADA$^*$ that is a extended version of StyleGAN-NADA~\cite{gal2021stylegan} to 3D generative model, we update  the EG3D model~\cite{chan2022efficient} using the directional CLIP loss as follows:
\small
\begin{align}
    \mathcal{L}_{\text{dir-CLIP}} = 1 - \frac{\langle \Delta I, \Delta T\rangle}{\|\Delta I\|\|\Delta T\|},
\end{align}
\normalsize
where \( \Delta I =  E_I^{{C}}(\xb^{\text{gen}}) - E_I^{{C}}(\xb^{\text{src}}), \Delta T =  E_T^{{C}}{(y^{\text{tar}}}) - E_T^{{C}}({y^{\text{src}}}) \).
We follow the instructions provided in the paper  and implement the loss and optimization part using the official StyleGAN-NADA codebase~\cite{gal2021stylegan}. To obtain the best results, we performed early stopping during total 2,000 iterations of model training.

In HyperDomainNet$^*$, a 3D extension of HyperDomainNet~\cite{alanov2022hyperdomainnet}, an indomain angle consistency loss is included in addition to the directional CLIP loss to maintain CLIP similarities between images before and after domain adaptation.
\small
\begin{align}
\mathcal{L}_{\text{indomain}} = \sum_{i, j}^n(\langle E^C_I(\xb^{\text{gen}}_i), E^C_I(\xb^{\text{gen}}_j)\rangle-\langle E^C_I(\xb^{\text{src}}_i), E^C_I(\xb^{\text{src}}_j)\rangle)^2,
\end{align}
\normalsize
We follow the instructions provided in the paper and implement the loss and optimization part using the official HyperDomainNet codebase~\cite{alanov2022hyperdomainnet}. To obtain the best results, we performed early stopping during total 2,000 iterations of model training.

StyleGANFusion~\cite{song2022diffusion} adopts the SDS loss~\cite{poole2022dreamfusion} to guide the text-guided adaptation of 2D and 3D generators using text-image-diffusion models, as defined:
\small
\begin{align}
    \mathcal{L}_{\text{SDS}} = \mathbb{E}_{t, \epsilon}[||\epsilonb_{\phi}(\xb_t,y, t) - \epsilonb||_2^2]
\end{align}
\normalsize
where $\xb$ is a perturbed image or latent thorugh forward diffusion, $\epsilonb \sim \mathcal{N}(0,1)$, $t \sim \mathcal{U}([1,T])$, and $T$ is the total diffusion steps.
We employ LPIPs regularization~\cite{zhang2018perceptual} and fine-tune the EG3D model~\cite{chan2022efficient} for 2,000 iterations following the instructions in the paper.

In DATID-3D~\cite{kim2022datid}, pose-aware target dataset is generated using Stable diffusion~\cite{rombach2022high}.
Then, the target dataset is refined through CLIP- and pose-reconstruction-based filtering process.
After that, EG3D model~\cite{chan2022efficient} is fine-tuned on the filtered dataset.
We faithfully follow the implementation details in the paper.
We generate 3,000 target images per target prompts and fine-tune the pretrained EG3D model~\cite{chan2022efficient} with a batch size of 20 making steps on 50k$\sim$100k training images.

\begin{table*}[!htb]
\centering
\caption{List of complete text prompts that corresponds to each concise prompt.}\label{tab2}%
\begin{adjustbox}{width=0.85\linewidth}
\begin{tabular}{ccc}
\toprule
Text types                   & Concise prompts  & Full text prompts                                                                          \\
 \midrule
\multirow{12}{*}{Characters} & SpongeBob        & \textit{"a FHD photo of a face of SpongeBob"}                                              \\
                             & Sesame Street    & \textit{"a FHD photo of a fluffy character from 'Sesame Street' "}                         \\
                             & Rango            & \textit{"a 3D render of a face of Rango from 'Rango'"}                                     \\
                             & Jack Skellington & \textit{"a FHD photo of a face of Jack Skellington from 'The Nightmare Before Christmas'"} \\
                             & Minions          & \textit{"a 3D render of Doraemon"}                                                         \\
                             & Homer Simpson    & \textit{"a 3D render of Homer simpson"}                                                    \\
                             & Doraemon         & \textit{"a 3D render of yellow Minion with two eyes from the 'Despicable Me' franchise"}   \\
                             & Sonic            & \textit{"a FHD photo of a face of Sonic the Hedgehog from 'Sonic the Hedgehog'"}           \\
                             & Shaun the Sheep  & \textit{"a FHD photo of Shaun the Sheep"}                                                  \\
                             & Teddy bear       & \textit{"a FHD photo of a Teddy bear"}                                                     \\
                             & Barbie doll      & \textit{"a FHD photo of a barbie doll"}                                                    \\
                             & Scooby-Doo       & \textit{"a 3D render of Scooby-Doo"}                                                       \\
\midrule
\multirow{9}{*}{Animals}     & Elephant         & \textit{"a FHD photo of a face of a Elephant"}                                             \\
                             & Turtle           & \textit{"a FHD photo of a face of a Turtle"}                                               \\
                             & Horse            & \textit{"a FHD photo of a face of a Horse"}                                                \\
                             & Iguana           & \textit{"a FHD photo of a face of a Iguana"}                                               \\
                             & Cameleon         & \textit{"a FHD photo of a face of a Cameleon"}                                             \\
                             & Bear             & \textit{"a FHD photo of a face of a Bear"}                                                 \\
                             & Dog              & \textit{"a FHD photo of a face of a Dog"}                                                  \\
                             & Goat             & \textit{"a FHD photo of a face of a Goat"}                                                 \\
                             & Cow              & \textit{"a FHD photo of a face of a Cow"}                                                  \\
\bottomrule
\end{tabular}
\end{adjustbox}
\end{table*}

\begin{table*}[!htb]
\centering
\caption{List of complete text prompts that correspond to each subclass prompt used for text-guided debiasing.}\label{tab3}%
\begin{adjustbox}{width=0.9\linewidth}
\begin{tabular}{ccc}
\toprule
Superclasses                              & Subclasses                & Subclasses prompts                                                             \\
\midrule
\multirow{4}{*}{Bear}                     & Black Bear                & \textit{"a FHD photo of a face of a Black Bear"}                               \\
                                          & Brown Bear                & \textit{"a FHD photo of a face of a Brown Bear"}                               \\
                                          & Giant Panda               & \textit{"a FHD photo of a face of a Giant Panda"}                              \\
                                          & Polar Bear                & \textit{"a FHD photo of a face of a Polar Bear"}                               \\
\midrule
\multirow{11}{*}{Dog}                     & Vizsla                    & \textit{"a FHD photo of a face of a Vizsla"}                                   \\
                                          & Pug dog                   & \textit{"a FHD photo of a face of a Pug dog"}                                  \\
                                          & German Shepherd           & \textit{"a FHD photo of a face of a German Shepherd"}                          \\
                                          & Golden Retriever          & \textit{"a FHD photo of a face of a Golden Retriever"}                         \\
                                          & Beagle                    & \textit{"a FHD photo of a face of a Beagle"}                                   \\
                                          & Rottweiler                & \textit{"a FHD photo of a face of a Rottweiler"}                               \\
                                          & Dachshund                 & \textit{"a FHD photo of a face of a Dachshund"}                                \\
                                          & Siberian Husky            & \textit{"a FHD photo of a face of a Siberian Husky"}                           \\
                                          & Schnauzer                 & \textit{"a FHD photo of a face of a Schnauzer"}                                \\
                                          & Cocker Spaniel            & \textit{"a FHD photo of a face of a Cocker Spaniel"}                           \\
                                          & Welsh Corgi               & \textit{"a FHD photo of a face of a Welsh Corgi"}                              \\
\midrule
\multirow{9}{*}{Zootopia}                 & Rabbit in Zootopia Style  & \textit{"a FHD photo of a Rabbit in Zootopia Style with big eyes"}             \\
                                          & Fox in Zootopia Style     & \textit{"a FHD photo of a Fox in Zootopia Style with big eyes"}                \\
                                          & Buffalo in Zootopia Style & \textit{"a FHD photo of a Buffalo in Zootopia Style with big eyes"}            \\
                                          & Cheetah in Zootopia Style & \textit{"a FHD photo of a Cheetah in Zootopia Style with big eyes"}            \\
                                          & Sheep in Zootopia Style   & \textit{"a FHD photo of a Sheep in Zootopia Style with big eyes"}              \\
                                          & Gazelle in Zootopia Style & \textit{"a FHD photo of a Gazelle in Zootopia Style with big eyes"}            \\
                                          & Tiger in Zootopia Style   & \textit{"a FHD photo of a Tiger in Zootopia Style with big eyes"}              \\
                                          & Bear in Zootopia Style    & \textit{"a FHD photo of a Bear in Zootopia Style with big eyes"}               \\
                                          & Koala in Zootopia Style   & \textit{"a FHD photo of a Koala in Zootopia Style with big eyes"}              \\
\midrule
\multirow{10}{*}{3D animation characters} & Toy Story                 & \textit{"a FHD photo of a character in film 'Toy Story' style"}                \\
                                          & Moana                     & \textit{"a FHD photo of a character in film 'Moana' style"}                    \\
                                          & How to Train Your Dragon  & \textit{"a FHD photo of a character in film 'How to Train Your Dragon' style"} \\
                                          & Brave                     & \textit{"a FHD photo of a character in film 'Brave' style"}                    \\
                                          & Coco                      & \textit{"a FHD photo of a character in film 'Coco' style"}                     \\
                                          & Ratatouille               & \textit{"a FHD photo of a character in film 'Ratatouille' style"}              \\
                                          & Rise of the Guardians     & \textit{"a FHD photo of a character in film 'Rise of the Guardians' style"}    \\
                                          & Tangled                   & \textit{"a FHD photo of a character in film 'Tangled' style"}                  \\
                                          & UP                        & \textit{"a FHD photo of a character in film 'UP' style"}                       \\
                                          & Moana                     & \textit{"a FHD photo of a character in film 'Moana' style"}                    \\
\bottomrule
\end{tabular}
\end{adjustbox}
\end{table*}

\subsection{User study}
To evaluate the quality of the generated samples and 3D shapes from the shifted generators from five different methods of text-guided domain adaptation for 3D generative models including our PODIA-3D, we conducted a user study using a survey platform. The study involved 60 participants, and a total of 10,500 votes were collected.
We applied each of the 5 methods of text-guided domain adaptation for 3D generative models, including our own method, to adapt the EG3D~\cite{chan2022efficient} generator to 7 text prompts, each converting a human face to a style with a large domain gap from the FFHQ~\cite{karras2019style} or AFHQ-cat~\cite{karras2021alias, choi2020stargan} domains, namely Horse', Cow', Elephant', Iguana', Turtle', Sesame street', `SpongeBob'.
For each text prompt, we sampled 30 images and a 3D shape from each generator, and placed the results of each method side-by-side.
To assist the participants in the user study, we included screenshots of Google search pages corresponding to each text prompt.
A total of 60 people completed the survey, providing 10,500 votes.
We asked the participants in the user study to rate the quality of the rendered 2D images on a scale of 1 to 5. They were presented with the following questions: 1) Do the rendered 2D images accurately reflect the semantics of the target text? (text-2D image correspondence), 2) Are the rendered 2D images realistic? (photorealism), and 3) Are the rendered 2D images diverse within the image group? (diversity). These questions are similar to those used in \cite{kim2022datid}.
Additionally, participants were asked to rate the accuracy of text-correspondence and the sense of depth and details of the 3D shapes on a scale of 1 to 5, based on the following questions: 4) Does the extracted 3D shape accurately reflect the semantics of the target text? (text-3D shape correspondence), and 5) Do the 3D shapes have a great sense of depth and detail? (sense of depth and details of 3D shapes). 
Finally, the mean score for each method was then calculated.

\section{Additional Results}
\label{supp_additional_results}
\subsection{Results of text-driven 3D domain adaptation}
We present additional results of qualitative comparison between our text-guided domain adaptation outcomes and other baselines in Fig~\ref{fig_supp2} and Fig~\ref{fig_supp3}. 
Our method empowers the EG3D~\cite{chan2022efficient} model to generate multi-view consistent images in a broad range of text-guided domains (animals, characters) with high text-image correspondence, diversity and excellent quality of 3D shape, while other baselines don't.

\subsection{SD vs DGD vs PPD}
Fig.~\ref{fig_supp4} presents additional comparison results of text-guided image-to-image translation and text-guided domain adaptation, evaluated using Stable diffusion (SD)\cite{rombach2022high}, the depth-guided diffusion (DGD)\cite{rombach2022high}, our pose-preserved diffusion (PPD) models, and specialized-to-general (S-to-G).

PPD enables high pose-consistency and text-image correspondence in image translation, which leads to successful domain adaptation of the EG3D~\cite{chan2022efficient} 3D generator while preserving 3D shapes.
Furthermore, S-to-G sampling help to address the issue of detail bias.

\section{Discussion}
\label{supp_additional_discussion}
\paragraph{Limitation.}
We investigate that domain adaptation to non-living objects, such as chairs and hamburgers, which lack directional information, leads to a low level of text-image correspondence as represented in Fig.~\ref{fig_supp5}.
Our text-guided domain adaptation process relies on the performance of text-to-image diffusion models, which means that any limitations of the chosen diffusion models are also present in our pipeline. 
For this work, we have utilized both the Stable Diffusion~\cite{rombach2022high} and Depth-Guided Diffusion~\cite{rombach2022high} models.
As stated in the Stable diffusion model card, the model has certain limitations that include the inability to achieve complete photorealism, compositionality, proper face generation, and the generation of images with languages other than English. These limitations may have an impact on the performance of our method.

\paragraph{Social impacts.}
Our novel PODIA-3D technique enables the creation of high-quality 3D samples in text-guided domains, even in cases where there are substantial gaps from the source domain. Importantly, this can be achieved without requiring any artistic skills. 
However, it is crucial to acknowledge that this technology has the potential to be misused for creating visuals that are upsetting or insulting.
As per the Stable diffusion model card~\cite{rombach2022high}, the model can be misused in various ways such as creating inaccurate, hurtful, or offensive depictions of individuals, and cultures.
Therefore, we strongly advise individuals to use our approach judiciously and solely for its intended purposes.

\end{document}